\pdfoutput=1

\documentclass[11pt]{article}

\usepackage[]{EMNLP2023}

\usepackage{times}
\usepackage{latexsym}

\usepackage[T1]{fontenc}

\usepackage[utf8]{inputenc}

\usepackage{microtype}

\usepackage{inconsolata}

\usepackage{graphicx}
\usepackage{latexsym}
\usepackage{amssymb}
\usepackage{amsmath}
\usepackage{amsthm}
\usepackage{enumitem}
\usepackage{graphicx}
\usepackage{color}
\usepackage{lscape}
\usepackage{booktabs}
\usepackage{tabularx} 
\usepackage{stackengine}
\usepackage{pifont}

\usepackage{hyperref}	
\usepackage{soul}
\usepackage[T1]{fontenc}
\usepackage[utf8]{inputenc}
\usepackage{float}
\usepackage{paralist}
\usepackage{subcaption}
\usepackage{titlesec}
\usepackage{inconsolata}
\usepackage{cleveref}
\usepackage{multirow}
\usepackage{mdframed}
\usepackage{anyfontsize}
\usepackage{makecell}
\usepackage{tcolorbox}

\definecolor{ForestGreen}{RGB}{34,139,34}
\newcommand{\cmark}{\textcolor{ForestGreen}{\ding{52}}}%
\newcommand{\xmark}{\textcolor{red}{\ding{55}}}%

\newcommand{\sayantan}[1]{\textcolor{black}{#1}}
\title{\textsc{Text2Afford}: Probing Object \underline{Afford}ance Prediction\\ abilities of Language Models solely from \underline{Text}}

  \author{Sayantan Adak, Daivik Agrawal, Animesh Mukherjee\thanks{*Equal advising} \and Somak Aditya\footnotemark[1] \\
  IIT, Kharagpur \\
  West Bengal -- 721302
}

\begin{document}
\maketitle
\begin{abstract}
We investigate the knowledge of object affordances in pre-trained language models (LMs) and pre-trained Vision-Language models (VLMs).
A growing body of literature shows that PTLMs fail inconsistently and non-intuitively, demonstrating a lack of reasoning and \textit{grounding}. To take a first step toward quantifying the effect of grounding (or lack thereof), we curate a novel and comprehensive dataset of object affordances -- \textsc{Text2Afford}, characterized by 15 affordance classes.  Unlike affordance datasets collected in vision and language domains, we annotate \textit{in-the-wild} sentences with objects and affordances. Experimental results reveal that PTLMs exhibit limited reasoning abilities when it comes to uncommon object affordances. We also observe that pre-trained VLMs do not necessarily capture object affordances effectively. Through few-shot fine-tuning, we demonstrate improvement in affordance knowledge in PTLMs and VLMs. Our research contributes a novel dataset for language grounding tasks, and presents insights into LM capabilities, advancing the understanding of object affordances. \footnote{Code and Data are available at \url{https://github.com/sayantan11995/Text2Afford}}
\end{abstract}

\section{Introduction}

Object affordance refers to the properties of an object that determine what actions a human can perform
on them \cite{plasticq}.
Gaining the knowledge of object affordances while learning textual representation from large corpora maybe hard; as in NLP, we lack corresponding images (or videos) which provides necessary visual cues such as shape, color, and texture to predict affordances. This lack of mapping or rather \textit{grounding} ability has been noted by many researchers in the context of large pretrained language models (PTLMs). Authors in \citet{bender-koller-2020-climbing} have pointed the lack of symbol grounding to be a fundamental factor behind PTLMs failing to grasp \textit{meaning} from \textit{form} (surface form text). The authors argue that language models which are exposed to only text (surface form) may never truly understand \textit{meaning}, as PTLMs are unaware of possible \textit{groundings} of the surface text. 
\begin{figure}[t]
    \centering
    \includegraphics[width=0.9\columnwidth]{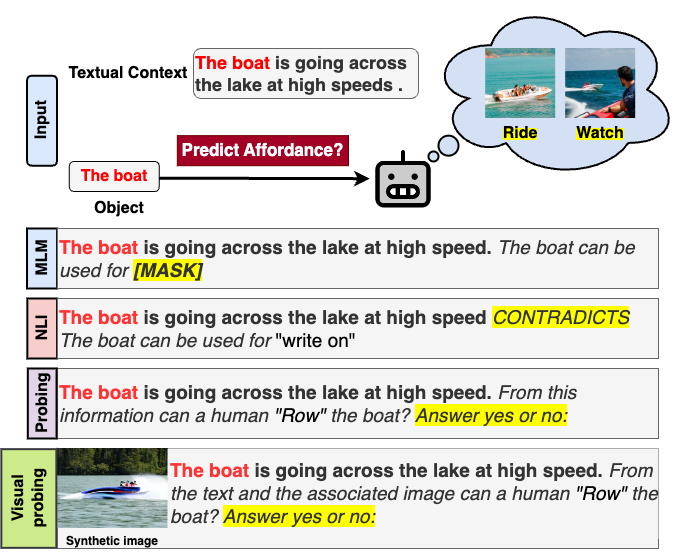}
    \caption{\footnotesize Overview of \textsc{Text2Afford} with its derived tasks.}
    
    \label{fig:introduction}
    \vspace{-1\baselineskip}
\end{figure}
Most current NLP datasets and tasks are not designed to evaluate \textit{grounding}, as it is hard to evaluate \textit{grounding} without any visual context. Here, we aim to \textit{quantify} the ability of pretrained models to learn \textit{affordances} -- which in turn requires the ability to \textit{ground} symbols in text to real-world objects.
 In other words, \textit{grounding} ability from text can enable understanding and reasoning about the physical properties of an object, which may help predict affordances. 

\begin{table*}[]\centering
\scriptsize
\renewcommand{\arraystretch}{1.2} 
\resizebox{\textwidth}{!}{
\begin{tabular}{
>{\centering\arraybackslash}m{15em}|
>{\centering\arraybackslash}m{5em}|
>{\centering\arraybackslash}m{5em}|
>{\centering\arraybackslash}m{5em}|
>{\centering\arraybackslash}m{15em}|
>{\centering\arraybackslash}m{0.12\textwidth}|
>{\centering\arraybackslash}m{0.08\textwidth}|
>{\centering\arraybackslash}m{0.08\textwidth}|
>{\centering\arraybackslash}m{0.08\textwidth}
}\Xhline{2\arrayrulewidth}
\rowcolor{gray!10} \textbf{Dataset} &\textbf{Train size} &\textbf{Dev size} &\textbf{Test size} &\textbf{Reasoning type} &\textbf{Source} &\textbf{Image-dependent} &\textbf{Targeted affordance} &\textbf{Publicly available}\\\hline
$\alpha$NLI \cite{Bhagavatula2020Abductive} &169,654 &- &1,532 &Abductive logical reasoning &Crowd-sourced &\xmark &\xmark &\cmark \\
$\alpha$ARCT \cite{niven-kao-2019-probing} &2420 &632 &888 &Abductive logical reasoning &Crowd-sourced &\xmark &\xmark &\cmark \\
FOLIO \cite{han2022folio} &1004 &204 &227 &Deductive logical reasoning &Expert written &\xmark &\xmark &\cmark \\
ANLI \cite{nie-etal-2020-adversarial} &162,865 &3,200 &3,200 &Deductive logical reasoning &Synthetic &\xmark &\xmark &\cmark \\
WinoLogic \cite{he-etal-2021-winologic} &- &- &562 &Deductive logical reasoning &Crowd-sourced &\xmark &\xmark &\cmark  \\
LogiQA \cite{10.5555/3491440.3491941} &7,376 &651 &651 &Mixed logical reasoning &Crowd-sourced &\xmark &\xmark &\cmark \\
LogiQA 2.0 \cite{10174688} &- &- &3,238 &Mixed logical reasoning &Crowd-sourced &\xmark &\xmark &\cmark \\\hline
PaCo \cite{qasemi-etal-2022-paco} &5,580 &1,860 &4,960 &Preconditioned commonsense &Crowd-sourced &\xmark &\xmark &\cmark \\
$\delta$-NLI\cite{rudinger-etal-2020-thinking} &36,999 &3,329 &3,512 &Defeasible commonsense &Other dataset &\xmark &\xmark &\cmark \\
WINOVENTI \cite{do-pavlick-2021-rotten} &- &- &4,352 &Commonsense with exceptions &Crowd-sourced &\xmark &\xmark &\cmark \\\hline
PVLIR \cite{qasemi2023preconditioned} & & &34,000 &Preconditioned visual commonsense &Other dataset &\cmark &\xmark &\xmark\\
Normlense \cite{han-etal-2023-reading}  &- &- &10,000 &Defeasible visual commonsense &Crowd-sourced &\cmark &\xmark &\cmark \\\hline
WinoViz \cite{jin2024winoviz} &- &- &1,380 &Reasoning object's visual property &Crowd-sourced &\xmark &\xmark &\xmark \\\hline
PROST \cite{aroca-ouellette-etal-2021-prost} &- &- &18,736 &Reasoning object's physical property &Other dataset &\xmark &\xmark &\cmark \\
NEWTON \cite{wang-etal-2023-newton} &- &- &2,800 &Reasoning object's physical property &Crowd-sourced &\xmark &\xmark &\cmark \\\hline
\citet{persiani-hellstrom-2019-unsupervised} &734,002 &- &314,572 &Object affordance without context &Synthetic &\xmark &\cmark &\xmark\\
\textbf{\textsc{Text2Afford} (Ours)} &- &- &35,520 (2368 * 15) &\textbf{Contextual object affordance} &\underline{Crowd-sourced} &\xmark &\cmark &\cmark\\
\hline
\end{tabular}
}
    \caption{\footnotesize Comparison of \textsc{Text2Afford} with other reasoning datasets.}

\label{tab:data_comparison}
\end{table*}

As another example, for the sentence ``an apple in the tree'', we should infer that the ``apple'' can be eaten, and is rollable. However we cannot roll an ``apple logo''. In computer vision and robotics efforts, an accompanying image (or video) often provides necessary information about shape and physical properties of the entity, which can be used to predict affordances \cite{zhu2014reasoning}. However such information is absent in NLP tasks. To capture this nuance, we annotate crowdsourced text intended for other tasks (such as NLI) with the objects and affordances. We use 15 affordance classes from \citet{zhu2014reasoning}. Through extensive pilot studies, we train a set of annotators using the \url{toloka.ai} platform. We choose $25$ highly-skilled annotators who annotated a total of $2368$ sentence-object pairs with 15 affordance classes, on a 0-3 Likert-like scale. For each sentence-object pair and each affordance class we ensure annotations from three annotators to enable majority votings. We name this novel dataset \textsc{Text2Afford}. 
We use the dataset for zero-shot evaluations of small LMs, open-source LLMs and also some VLMs by forming different task setups. Figure~\ref{fig:introduction} presents an example from \textsc{Text2Afford} and the derived tasks (detailed in Section~\ref{sec:task_description}). We evaluate the effect of few-shot fine-tuning on few PTLMs and VLM. Our contributions can be summarized as follows.

\noindent
$\bullet$ We curate a novel large scale crowdsourced text to affordance dataset -- \textsc{Text2Afford}, consisting of 35,520 test data points (2368 sentence-object pairs with 15 unique affordance classes per pair). We ensure at least three annotations for each sentence-object pair for each class. 

\noindent
$\bullet$ \sayantan{Using \textsc{Text2Afford}, we perform zero-shot evaluation of several state-of-the-art PTLMs along with a few VLMs in different settings to identify the extent to which they gain the knowledge of affordance during pretraining. We further ensemble the VLM and the PTLM predictions to examine whether pre-training with images can enrich affordance prediction from text. Overall, we observe that the SOTA LLMs face difficulties predicting contextual object affordances solely from text (accuracy $< 55\%$) and the performance gets slightly enhanced when using powerful VLMs in presence of synthetic images. } 

\noindent
$\bullet$ \sayantan{We also fine-tune few PTLMs on a small subset of our data as well as on some commonsense reasoning tasks to understand how quickly the affordance knowledge get scaled up and how far the affordances are related to commonsense knowledge. In addition, we examine the in-context learning (ICL) ability of few of the SOTA generative LLMs and VLMs in affordance prediction task. We find that the pre-trained encoder based models gain some knowledge about object affordance during fine-tuning using the commonsense reasoning dataset.}

\noindent
$\bullet$ Additionally through finetuning on our dataset,  we show that knowledge of affordance can improve model's physical reasoning capability. 


\section{Related work}

\noindent\textbf{Reasoning about object affordances.}  Object affordances has been extensively studied in Computer Vision and Robotics~(\citet{sun2014object,zhu2014reasoning}).
Recent methods employ deep learning approaches to detect object affordance. 
\citet{nguyen2017object} applies an object detector, CNN and dense conditional random fields to detect object affordance from RGB images. 
\citet{persiani-hellstrom-2019-unsupervised} extracts object-action pairs from web corpora using semantic role labelling. 
In contrast, we propose a crowd-sourced text only affordance dataset to audit the strength of SOTA LLMs and VLMs to reason about contextualized object affordance.  

\noindent\textbf{Probing methods.} \citet{talmor2020olmpics} utilizes probing and employs Multi-choice MLM (Masked Language Modelling) and Multi-choice QA (Question Answering) setup to capture reasoning capabilities of pre-trained Language Models. 
\citet{yang-etal-2022-z} examines zero-shot prediction performances on different tasks by LLM through novel visual imagination. 
\citet{aroca-ouellette-etal-2021-prost} highlights the shortcomings of state-of-the-art pre-trained models in physical reasoning, with a further performance decline observed when incorporating option shuffling and superlatives in reasoning questions.
\citet{liu2022things} proposes a novel spatial commonsense probing
framework to investigate object scales and positional relationship knowledge in text-based pre-trained models and models with visual signals. \citet{joshi2020taxinli} uses probing methods to investigate a more fine-grained logical reasoning capabilities of pre-trained models. 

\noindent\textbf{Reasoning tasks and dataset.}~
Reasoning about object affordance require a sort of commonsense reasoning. 
A series of works (\citet{singh-etal-2021-com2sense}, \citet{srivastava2023beyond}, \citet{Bisk2019PIQARA}, \citet{huang-etal-2019-cosmos}, \citet{talmor-etal-2019-commonsenseqa}, \citet{talmor2021commonsenseqa}, \citet{zellers-etal-2018-swag}) study the text based commonsense knowledge of language models. 
Dataset such as $\delta$-NLI~\cite{rudinger-etal-2020-thinking} focuses on defeasible inference of commonsense knowledge; PaCo~\cite{qasemi-etal-2022-paco} and PInKS~\cite{qasemi-etal-2022-pinks} deal with preconditioned commonsense inference of language models. PVLIR~\cite{qasemi2023preconditioned}, Normlense~\cite{han-etal-2023-reading} use images as precoditions to reason about defeasible commonsense norms. However, none of these specifically focus on reasoning of object affordance. \citet{wang-etal-2023-newton} proposes a benchmark of object-attribute pairs plus a diverse set of questions to reason object's physical properties. \citet{aroca-ouellette-etal-2021-prost} tackles physical and affordance reasoning from an object-centric approach. \citet{persiani-hellstrom-2019-unsupervised} attempts to extract common object-action pairs from web corpora. In Table~\ref{tab:data_comparison}, we demonstrate the comparison of \textsc{Text2Afford} with other datasets which perform different kind of reasoning tasks. We emphasize that, \textsc{Text2Afford} is the largest crowdsourced publicly available text based contextualized affordance dataset with a test size of 35,520 (2368 sentence-object pairs and 15 affordance classes).\\
\noindent\textbf{Present work.}
Although a substantial number of work study the reasoning capabilities of language models and propose commonsense reasoning datasets, however, none of these work concentrate specifically on evaluating the knowledge of affordance and contextual affordance prediction \textit{solely} from text. To bridge this gap, we present a reliable crowdsourced test dataset for identifying the contextualized affordance prediction capability of LLMs as well as VLMs. Our results show that the advanced large language models fail to understand an object’s physical properties aka the affordances from texts, and there is significant room for improvement which may further motivate researchers to explore models that explicitly learns to \textit{ground} objects in text to predict its physical properties and affordances.

\begin{table*}[!htp]
\centering
\scriptsize
\renewcommand{\arraystretch}{1.1} 
\resizebox{\textwidth}{!}{
\begin{tabular}{
    >{\centering\arraybackslash}m{2.2cm}|
    >{\centering\arraybackslash}m{4cm}|
    >{\centering\arraybackslash}m{6cm}|
    >{\centering\arraybackslash}m{7cm}
}
\Xhline{2\arrayrulewidth}
\rowcolor{gray!20}
\textbf{Agreement category} & \textbf{Affordance classes} & \textbf{Objects} & \textbf{Object-affordance pair}\\\hline
\end{tabular}
}
\renewcommand{\arraystretch}{1.1} 
\resizebox{\textwidth}{!}{
\begin{tabular}{
    >{\raggedright\arraybackslash}m{2.2cm}|
    >{\raggedright\arraybackslash}m{4cm}|
    >{\raggedright\arraybackslash}m{6cm}|
    >{\raggedright\arraybackslash}m{7cm}
}
\rowcolor{green!15} High agreement (\textbf{>0.75}) & Row, Feed, Ride, Fix & the horse, striped white shirts, a brown paper sack, Chinese lanterns, Adrin's sword, The movie & breakfast-Feed, a horse-Watch, crops-Fix, sports-Grasp, sports-Lift, sports-Push, the phone-Feed, football-Ride\\
\hline
\rowcolor{yellow!10} Medium agreement (\textbf{>0.4 \& <0.75}) & Throw, PourFrom, WriteWith, LookThrough, Lift, Grasp, Play, Push & A red flag, An arrow, Art, Automatic weapons, Babies, Black-and-white TV & computers-WriteWith, cats-Feed, football-Play, book-WriteWith, the door-Push\\
\hline
\rowcolor{red!15} Low agreement (\textbf{<0.4}) & Watch, SitOn, TypeOn & Brandy from Spain, stone circles, iron, batteries, his fist, historical artifact, gift, olive oil, outdoor tables, bumper sticker on a car & weapon-Push, The table-Lift, boat-Fix, paintings-LookThrough, cats-Throw\\
\hline
\end{tabular}
}
\caption{\footnotesize Agreement based on difficulty in disambiguating different affordance classes, objects and object-affordance pairs.}
\label{tab:classwise_agreement}
\end{table*}

\section{\textsc{Text2Afford} dataset construction}

We select $20,000$ sentences from a crowdsourced English dataset (XNLI English) \cite{conneau2018xnli}\footnote{We choose XNLI as a source to facilitate multilingual extensions of our dataset.} and extract the noun phrases using the Stanford CoreNLP tool. As we restrict to the affordances that humans can directly perform, we filter the phrases which do not represent a tangible object (using ConceptNet). We manually filter out objects that cannot be acted upon directly by humans (such as school, building). After this preprocessing, we obtain a set of sentence-object pairs ($\langle x_i, o_i\rangle$), where the sentence acts as the context for the corresponding object. Each sentence on average has 2-3 such objects. 
We use the 15 predefined affordance classes from \citet{zhu2014reasoning} to label each sentence-object pair for annotation. \\
\noindent We utilize the Toloka platform\footnote{\url{https://toloka.ai/}} for conducting the data annotation. We design an interface on this platform, containing clear instructions and examples for annotating the data. We conduct two rounds of pilot studies along with additional AMA (Ask Me Anything) sessions to analyze the subjective understanding of the annotators and, thereby, only select the high quality, serious annotators. A total of 114 annotators participated in the pilot study, and out of that we finally engage 25 skilled annotators to annotate a total of \underline{2,368 sentence-object pairs} each containing 15 affordance classes. Each datapoint (i.e., sentence-object pair along with an affordance class) has been annotated by \textit{three} different annotators. We provide the details of the \textit{pilot studies \& annotator training} in Appendix~\ref{appendix:dataset_construction_details}. By evaluating the complexity of the task for the annotators from the pilot studies, we intentionally consider a relatively small number of datapoints at a point for the annotation. This leads us to a total of \underline{10 phases} to complete the final annotation. We carefully reviewed each annotation and provided feedback with guidance in case of mistakes. For instance, annotators initially got confused with the affordance `Watch' as human can \textit{watch} any visual objects. In another instance, some annotators asked whether `Throw' can be valid affordance for the object `Kittens' as humans can perform `Lift', `Throw' to the object `Kitten'. We discussed these types of ambiguities with the annotators after each phase. Throughout each of the data annotation phases, we put scrupulous attention to quality control, including iterative annotation refinement, and manual evaluation. The overall statistics for this \textit{currently} constructed dataset -- \textsc{Text2Afford} is in Table~\ref{tab:data_stat}. The \textsc{Text2Afford} dataset consists of $2368$ sentence-object pairs having  $\sim 100k$ annotations (2368 $\times$ 15 $\times$ 3). For further details of the dataset construction, and our method of handling ambiguous scenarios, we refer the reader to Appendix~\ref{appendix:dataset_construction_details}.

\begin{table}[!t]\centering
\resizebox{\columnwidth}{!}
{%
\begin{tabular}{l|l}\Xhline{2\arrayrulewidth}
\# of sentence-object pairs annotated &2368 \\
\# of affordance class &15 \\
\# of instances annotated &106560 (2368 $\times$ 15 $\times$ 3) \\
Avg \# of objects / class &333 \\
Most prominent class &Lift (851 objects) \\
Least prominent class &WriteWith (3 objects) \\
Total skilled annotators used &25 \\
Avg agreement (Kripendorff's $\alpha$
) &0.68 \\
\hline
\end{tabular}%
}
\caption{\footnotesize \textsc{Text2Afford} dataset statistics. \# of instances annotated: (\# of <s-o> pairs) * (\# of classes) * (\# of annotations per class).}

\label{tab:data_stat}
\vspace{-1\baselineskip}
\end{table}

\begin{figure}[htbp]
    \centering
    \includegraphics[width=\columnwidth]{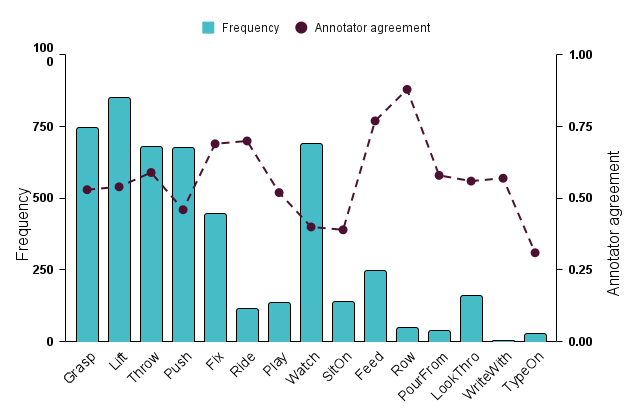}
    \caption{\footnotesize Classwise distribution of the number of objects and the annotator agreement.}

\label{fig:affordance_class_distribution_normal}
\end{figure}

\noindent\textbf{\textsc{Text2Afford} data exploration.} We observe that classes such as `Grasp', `Lift', `Throw', `Push', and `Watch' are the most common affordances for the objects present in the dataset (see Figure~\ref{fig:affordance_class_distribution_normal}). Most frequent objects and their corresponding agreement scores are shown in Appendix~\ref{subsec:most_frequent_objects} Fig.~\ref{fig:frequent_object_combined}. We observe, agreement scores are fairly uniform ($0.5$-$0.6$) for frequent objects, with high agreement for some frequent objects ($0.8$ for ``the movie''). In Figure~\ref{fig:affordance_class_correlation} (see Appendix~\ref{subsec:corr_afford}), we also see that `Grasp', `Lift', and `Throw' are highly correlated classes. There is similar positive correlation between the class `SitOn' and `Ride', and some correlation between `Watch' and `LookThrough'.\\
In Table~\ref{tab:classwise_agreement}, we list down the affordance classes based on the annotator agreement score, and divide it into \textit{three} categories to understand which of the affordance classes pose the most and least difficulties for the human annotators. We observe that the classes - `Watch', `SitOn', and `TypeOn' are the most difficult to disambiguate. Further, to explore the difficulty of understanding contextual object affordance, we employ three \textit{na\"ive annotators} to annotate some samples of the \textsc{Text2Afford}, and we observe that on an average in \underline{88\%} cases the humans are able to predict affordance correctly, and in some cases the \textit{context} introducing inherent difficulty for predicting affordance. Details of the study is provided in Appendix~\ref{appendix:additional_data_analysis}.



\section{Task description}
\label{sec:task_description}
\sayantan{Our first objective is to \textit{audit} the strength of Large Language Models in identifying the pre-defined affordance classes of objects from text in zero-shot settings. Given a textual context, and the object, the task is to predict whether a particular affordance class is applicable to that object conditioned on the context. We majorly leverage 4 types of task setup for the experiments. For the encoder based models (e.g., RoBERTa, BERT) we choose Masked Language Modelling (MLM) and Natural Language Inference (NLI) based setup, and for the generative models we adopt 2 types of probing setup (text only, text+image) to formalize the task. Table~\ref{tab:tasks} demonstrates different types of tasks that we engage for conducting the experiments from the \textsc{Text2Afford} dataset. 
}

\begin{table*}[]
\centering
\scriptsize
\renewcommand{\arraystretch}{1.1}
\resizebox{\textwidth}{!}{
\begin{tabular}{
    >{\centering\arraybackslash}m{0.15\textwidth}|
    >{\centering\arraybackslash}m{0.15\textwidth}|
    >{\centering\arraybackslash}m{0.55\textwidth}|
    >{\centering\arraybackslash}m{0.3\textwidth}
}
\Xhline{2\arrayrulewidth} 
\rowcolor{gray!10} \textbf{Model architecture} &\textbf{Tasks} &\textbf{Input instance} & \textbf{Output} \\
\end{tabular}
}
\resizebox{\textwidth}{!}{
\begin{tabular}{
    >{\centering\arraybackslash}m{0.15\textwidth}|
    >{\centering\arraybackslash}m{0.15\textwidth}|
    >{\raggedright\arraybackslash}m{0.55\textwidth}|
    >{\raggedright\arraybackslash}m{0.3\textwidth}
}
\hline 
\multirow{3}{*}{Encoder based} &MLM & All the women in India wear bangles [SEP\_TOKEN] bangles can be used for [MASK\_TOKEN] by human &Probabilities of each affordance classes as the [MASK\_TOKEN]\\\cline{2-4}
&\multirow{2}[1]{*}{NLI} &{\underline{premise}: All the women in India wear bangles} &  \multirow{2}[1]{*}{Entailment scores for each affordance classes} \\
& & {\underline{hypothesis}: bangles can be used for <Affordance> by human} \vspace{0.5em} & \\\cline{1-4}
\multirow{2}[8]{*}{Generation based} &Probing with text  & Consider the sentence - 'All the women in India wear bangles'. Now, from this information can a human <Affordance> bangles? Answer YES or NO: &YES\textbackslash NO \\\cline{2-4}
& Probing with text and image & \begin{minipage}{0.52\textwidth}
    \vspace{0.25em} \includegraphics[width=0.18\linewidth, height=0.15\linewidth]{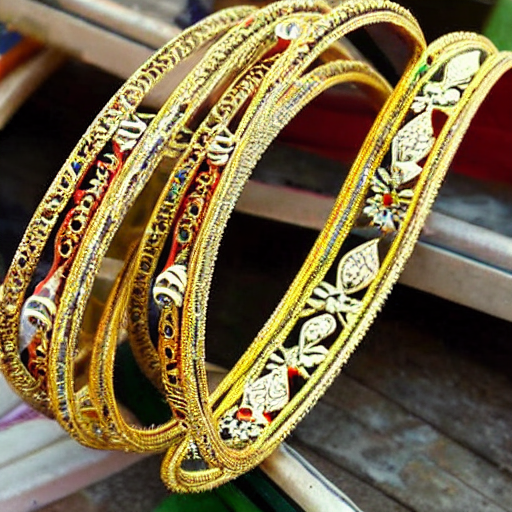} \hfill 
    \begin{minipage}[b]{0.8\linewidth} 
        Consider the sentence - 'All the women in India wear bangles'. Now, from this information can a human <Affordance> bangles? Accompanying this query is an image of the bangles. Answer YES or NO:
    \end{minipage}
  \end{minipage} \vspace{0.25em} &YES\textbackslash NO \\
\hline
\end{tabular}
}
\vspace{-0.3\baselineskip}
\caption{\footnotesize Overview of the tasks using \textsc{Text2Afford}. For detailed prompting see Appendix~\ref{tab:prompts}.}
\label{tab:tasks}
\vspace{-1\baselineskip}
\end{table*}

\section{Experiments}
We explore various state-of-the-art baselines using pre-trained language models (RoBERTa-large, BART-large), instruction-fine-tuned large language models (e.g., FLAN-T5, Falcon, ChatGPT, Llama-3), pre-trained multi-modal vision and language architectures (CLIP-ViT, ViLT, InstructBLIP, IDEFICS, LLaVA). We observe whether these models gain the knowledge of affordances through their pre-training, fine-tuning on commonsense tasks (NLI, PIQA), or few-shot fine-tuning scenarios. 

\subsection{Zero-shot affordance prediction}
\subsubsection{Pre-trained language models}
We frame the zero-shot prediction task in different ways.\\
\noindent\textbf{MLM based approach.} Here, we pose the zero-shot task as masked word prediction problem. We choose BERT-large-uncased, RoBERTa-large \cite{zhuang-etal-2021-robustly}, and BART-large \cite{lewis2020bart} models for the experiment.
We pass the sentence and prompt separated by a \texttt{[SEP]} token as an input to the model. We use the prompt  ``\texttt{<Object> can be used for <MASK\_TOKEN> by human}'' and obtain the probability of each affordance label using the logit corresponding to the \texttt{<MASK\_TOKEN>}.\\ 
\noindent\textbf{Predictions from generative LLMs.} \label{exp:generative_LLM} We pose the task as `YES\textbackslash NO' questions-answering format and apply autoregressive language models such as FLAN-T5 \cite{chung2022scaling} (large, xl, and xxl), Falcon \cite{almazrouei2023falcon} (7b and 40b), Llama-3 \footnote{\url{https://github.com/meta-llama/llama3}}, ChatGPT to get the predictions. 
We provide with a `YES\textbackslash NO' question-answer based prompt to the LLMs to predict whether a particular affordance can be performed on the given object. Based on rigorous prompt engineering we choose specific prompts for the different models as shown in the Appendix Table~\ref{tab:prompts}. We map `YES\textbackslash NO' predictions to 1\textbackslash 0 labels respectively.

\subsubsection{Commonsense reasoning tasks}
To understand whether the injection of the common
sense knowledge in the pre-trained models can enhance the performance of the affordance prediction, we first fine-tune the pre-trained models on common
sense reasoning dataset such as PIQA \cite{Bisk2019PIQARA}. Then we run the fine-tuned models on our dataset using the MLM setup. We use BERT-base, BERT-large, RoBERTa-large, and BART-large models.\\ 
Apart from this, we leverage RoBERTa-large and BART-large fine-tuned on the Multi-genre NLI (MNLI) corpus \cite{williams-etal-2018-broad} to evaluate on NLI setup. We utilize the sentence as premise and use the hypothesis as ``\textit{<object> can be used for <affordance> by human}'' for each object-affordance pair, and use the entailment score to rank the affordance classes and report mAP and accuracy. Details of the experiment can be found in Appendix~\ref{appendix:zero-shot-commonsense-reasoning}.

\subsubsection{Multimodal models}
\label{sec:multimodal}
We explore both unimodal and multi-modal task setup for pre-trained vision and language models. 

\subsubsection*{Text-only MLM setup}
VLMs are pre-trained on large datasets having both image and text. The main goal of their pre-training is to capture some visual knowledge corresponding to the text while pre-training on multi-modal dataset such as image-caption pairs. To examine this, we first use the vision-language model CLIP, by providing only text prompt as the input and predict the affordance in an MLM setup.
\subsubsection*{Multimodal task setup}
Images contain necessary information about shape, texture, and size of objects that can be utilized to effectively predict an object affordance (such as the handle of the bucket can be used to grasp and lift). Hence, we also convert the problem into a multi-modal task by synthesizing corresponding images from the context sentence, and  predict the affordance of an object (mentioned in the sentence) based on the input.\\ 
\noindent\textbf{Synthesizing images.}  In this setup, we use two different techniques to synthesize \textit{semantically close} images to corresponding context sentences using 1) retrieval and 2) generation. We further use top five images for both, to get an accurate estimation. \\
\noindent\textit{Image retrieval}: We use the CLIP \cite{radford2021learning} based sentence-transformers architecture to search for top five semantically similar images for each of the contexts from the Visualgenome \cite{krishna2017visual} dataset.\\
\noindent\textit{Image generation}:  We adopt the generative \textit{StableDiffusion} \cite{rombach2022high} model to generate top five images based on the sentence as a text prompt. Details can be found in the Appendix~\ref{appendix:multimodal-task-setup}. 

\noindent We use the top five retrieved images by using retrieval and generation methods each. We use \textit{CLIP} \cite{radford2021learning} and \textit{ViLT} \cite{kim2021vilt} as our vision-text models.  CLIP has a text encoder $f_T$ and a visual encoder $f_V$, which can project text and image into the shared latent space. We aggregate the $k$ (=5) corresponding images and use CLIP to compute the relevance score of (x, y): $Score_{VI}(x, y) = \frac{1}{k}  \sum_{i=1}^{K}{\cos{(f_T(x), f_v(I^k_y))}}$,
where $I^k_y$ is the $k^\textrm{th}$ image for the input text $y$. 
In the ViLT model, we provide the text prompt along with the representative images as input to predict the masked token. We use the same prompt as the previous MLM task (i.e., ``\textit{<Object> can be used for <MASK\_TOKEN> by human.}'') and get the probability of each affordance class as the logit corresponding to the <MASK\_TOKEN>.\\
\noindent\textbf{Text generation based. } Similar to section~\ref{exp:generative_LLM}, we utilize state-of-the-art VLMs to make predictions regarding object affordances. We provide with a `YES\textbackslash NO` question answering based text prompt along with the aligned images as input to the VLMs, and the model should generate an answer whether a particular affordance can be performed on the given object. We use state-of-the-art VLMs such as IDEFICS \cite{laurenccon2023obelisc}, LLaVA \cite{liu2023llava}, InstructBLIP \cite{instructblip} for this task. The text prompt used for the models can be found in the Appendix~\ref{appendix:prompts}, Table~\ref{tab:prompts}.

\paragraph{Ensemble language and vision prediction.} Following \citet{yang-etal-2022-z}, we use the weighted sum as the late fusion over the final output probabilities of each affordance class from the language and multi-modal models. Experimental details can be found in Appendix~\ref{appendix:exp_ensemble_VL}. 

\subsection{Few-shot prediction}
We conduct few-shot experiments by 1) fine-tuning the encoder based models, 2) randomly selecting 5 demonstration examples for the generative models to perform few-shot in-context learning (ICL). We consider the 62 annotated objects and corresponding 15 affordance classes by \citet{zhu2014reasoning} for the few-shot based experiments. \\
\noindent\textbf{Training data} To create few-shot training examples for fine-tuning encoder based PTLMs, we take all the 62 objects, and for each object we randomly select exactly 1 positive affordance class (i.e., the class label annotated as 1) and 1 negative affordance class (i.e., the class label annotated as 0) for generating the training prompt. Overall they constitute 124 training examples (62 sentence-object pairs and 2 selected classes for each) for the few-shot experiment. For more details of the training data curation and the selection of examples for in-context learning, refer to Appendix~\ref{appendix:few-shot-training-data}\\
\noindent\textbf{Experimental setup.} We fine-tune the encoder based language models using the training data, and for the generative LLMs and the VLMs, we utilize the training data to select in-context demonstration examples.\\
\noindent\textit{Fine-tuning PTLM:} We fine-tune the encoder based PTLMs in NLI based setup having the context sentence as premise and use same hypothesis (i.e., ``<object> can be used for <affordance> by human'') which we use in the zero-shot settings. We use BERT-large-uncased, RoBERTa-large and BART-large for fine-tuning in this setup. For implementation details refer to Appendix~\ref{appendix:few-shot-experimental_setup}\\
\noindent\textit{In-context learning for generative models}: We employ the same generative LLMs as well as VLMs to perform affordance prediction using \textit{five} demonstration examples from the training data. We use the same text prompt as zero-shot setting and concatenate the five demonstration examples along with corresponding label (i.e., `NO' for positive class, and `NO' for the negative class) to the prompt and ask the LLMs and VLMs to predict the affordance. In case of the VLMs, we do not provide any additional image example here.
\vspace{-0.3\baselineskip}

\section{Benchmarking \textsc{Text2Afford} prediction}
\paragraph{Evaluation metric.} To assess the performance of the zero-shot affordance prediction, we calculate accuracy in the following way. Each affordance class is treated as a binary classification problem, where a value of 1 represents a positive class indicating that the affordance can be performed on the object, and a value of 0 represents a negative class indicating that the affordance cannot be performed.\\
For each positive class $ \in \{P_1, P_2,.. P_n\}$, we compare the predicted scores of that affordance class with the predicted scores of the negative classes $ \in \{N_1, N_2,.. N_m \}$. If the predicted score of the positive class is higher than the predicted score of all the negative classes (i.e., $p(P_i) > p(N_j)_{\forall j}$), we increment the correct count by 1\footnote{During calculation we discard the cases when there is no positive class for a sentence-object pair in the ground truth. We do not find any instance where no negative class is present.}. Conversely, if the predicted score of the negative class is higher, we increment the wrong count by 1.
The final accuracy is calculated by dividing the total number of correct counts by the total number of the instances. To rank the affordance classes based on the predicted score, we also report the Mean Average Precision (mAP@$K$, where $K$ is the number of affordance classes). 

\begin{table}[!htb]
\scriptsize
\centering
\renewcommand{\arraystretch}{1.1}
\resizebox{\columnwidth}{!}{
\begin{tabular}{l|
>{\centering\arraybackslash}m{3em}
>{\centering\arraybackslash}m{3em}|
>{\centering\arraybackslash}m{3em}
>{\centering\arraybackslash}m{3em}|
>{\centering\arraybackslash}m{3em}
>{\centering\arraybackslash}m{3em}|
>{\centering\arraybackslash}m{3em}
>{\centering\arraybackslash}m{3em}
}\Xhline{2\arrayrulewidth}
\rowcolor{gray!10} \multicolumn{9}{c}{\textbf{Encoder based}} \\\hline
\rowcolor{red!10} \multicolumn{9}{c}{\textbf{NLI based}} \\\hline
\multirow{2}{*}{\textbf{Model}}&\multicolumn{2}{c|}{\textbf{Actual}} &\multicolumn{2}{c|}{\textbf{Fine-tuned}} & \multicolumn{2}{c|}{\textbf{LM + VI (CLIP)}} &\multicolumn{2}{c}{\textbf{LM + VI (ViLT)}} \\\cline{2-9}
 &\textbf{Acc} &\textbf{mAP} &\textbf{Acc} &\textbf{mAP} &\textbf{Acc} &\textbf{mAP} &\textbf{Acc} &\textbf{mAP} \\\hline
RoBERTa-large-mnli &0.64 &0.43 &0.72 &0.49 &\textbf{0.79} &0.52 &\textbf{0.79} &\textbf{0.54} \\
BART-large-mnli &0.65 &0.38 &0.69 & 0.48 &0.62 &0.4 &0.64 &0.43 \\\hline
\rowcolor{blue!10}\multicolumn{9}{c}{\textbf{MLM based}} \\\hline
BERT-large-uncased &0.46 &0.26 &0.58 &0.33 &0.55 &0.38 &0.53 &0.37 \\
RoBERTa-large &0.55 &0.36  &\textbf{0.77} &\textbf{0.49} &0.61 &0.41 &0.62 &0.43 \\
BART-large &0.47 &0.28 &0.65 &0.38 &0.56 &0.35 &0.52  &0.34 \\\hline
\rowcolor{brown!20} \multicolumn{9}{c}{\textbf{Multi-modal models (zero-shot)}}\\\hline
CLIP-VIT (text-only) &0.47 &0.34 &- &- &- &- &- &- \\
CLIP-VIT (retrieval) &0.56 &0.35 &- &- &- &- &- &- \\
CLIP-VIT (generation) &\textbf{0.61} &\textbf{0.4} &- &- &- &- &- &-\\
ViLT (retrieval) &0.41 &0.31 &- &- &- &- &- &-\\
ViLT (generation) &0.44 &0.32 &- &- &- &- &- &-\\
\hline
\end{tabular}
}
\caption{\footnotesize Performance for affordance prediction using encoder based models. Acc: Accuracy, LM: Language model, VI: Vision. Only LMs are ensembled with VI. The best results are in bold.}\label{tab:zero_shot_affordance_prediction}
\end{table}

\begin{table}[!htp]\centering
\scriptsize
\renewcommand{\arraystretch}{1.1} 
\resizebox{\columnwidth}{!}{
\begin{tabular}{l|cc|cc}\Xhline{2\arrayrulewidth}
\rowcolor{gray!10} \multicolumn{5}{c}{\textbf{Generation based}} \\\hline
\rowcolor{yellow!20} \multicolumn{5}{c}{\textbf{Predictions from generative LLM}} \\\hline
\textbf{Model} &\multicolumn{2}{c}{\textbf{Acc (zero-shot)}} &\multicolumn{2}{c}{\textbf{Acc (ICL)}}\\\hline
Random baseline &\multicolumn{2}{c|}{0.18} &\multicolumn{2}{c}{-}\\
FLAN-T5-large &\multicolumn{2}{c|}{0.06} &\multicolumn{2}{c}{0.13{$\pm0.04$}}\\
FLAN-T5-xl &\multicolumn{2}{c|}{0.07} &\multicolumn{2}{c}{0.21{$\pm0.03$}}\\
FLAN-T5-xxl &\multicolumn{2}{c|}{0.33} &\multicolumn{2}{c}{0.39{$\pm 0.04$}}\\
Falcon-7b-instruct &\multicolumn{2}{c|}{0.19} &\multicolumn{2}{c}{0.24{$\pm 0.03$}}\\
Falcon-40b-instruct &\multicolumn{2}{c|}{\textbf{0.43}} &\multicolumn{2}{c}{\textbf{0.47{$\pm 0.06$}}}\\
Llama-3-8b-instruct &\multicolumn{2}{c|}{0.36} &\multicolumn{2}{c}{0.43{$\pm 0.05$}}\\
ChatGPT (GPT-3.5 turbo) &\multicolumn{2}{c|}{0.41} &\multicolumn{2}{c}{0.44{$\pm 0.05$}} \\\hline
\rowcolor{brown!20} \multicolumn{5}{c}{\textbf{Multi-modal models}} \\\hline
\multirow{2}{*}{\textbf{Model}} &\multicolumn{2}{c|}{\textbf{Acc (zero-shot)}} &\multicolumn{2}{c}{\textbf{Acc (ICL)}}\\
&\textbf{IR based} &\textbf{IG based} &\textbf{IR based} &\textbf{IG based}\\\hline
Idefics-9b-instruct &0.26 &0.25 &0.36{$\pm0.02$} &0.37{$\pm0.03$}\\
Llava-1.5-7b &0.32 &0.34 &0.36{$\pm0.03$} &0.40{$\pm0.04$}\\
InstructBlip-vicuna-13b &0.37 &0.39 &0.43{$\pm0.03$} &0.45{$\pm0.03$}\\
InstructBlip-flan-t5-xl &0.12 &0.16 &0.15{$\pm0.02$} &0.18{$\pm0.02$}\\
InstructBlip-flan-t5-xxl &0.39 &\textbf{0.45} &0.48{$\pm0.04$} &\textbf{0.53{$\pm0.05$}}\\
\hline
\end{tabular}
}
\caption{\footnotesize Zero-shot and in-context learning (ICL) performance for affordance prediction using generative models. IR: Image Retrieval; IG: Image Generation. Number of demonstration examples used for ICL = 5. We also mention the variance over different selections of examples. The best results are in bold.}

\label{tab:affordance_prediction_generative}
\end{table}

\noindent\textbf{Zero-shot performance.} Table~\ref{tab:zero_shot_affordance_prediction} shows the results of the zero-shot affordance predictions from the mentioned models. The second column (i.e., Actual) indicates the values from the original LM and multi-modal models. The third and fourth columns (i.e., LM + VI) indicate the performances of ensembling language models with two of the multi-modal models we used. We observe that, the PTLMs have some knowledge about object affordances, but they still lack the comprehensive reasoning ability about these affordances, which is reflected in the low mAP values. Further, the performances vary across different settings. In case of NLI based setup, the fine-tuned RoBERTa and BART models show improvement in the performance, which indicates that \textit{during fine-tuning on MNLI dataset, those models gain some reasoning ability}. In Table~\ref{tab:affordance_prediction_generative} we show the generation based results in a zero-shot setting. In case of FLAN-T5-large model, where we use it to predict a binary label (YES\textbackslash NO) for an affordance class, the performance drops significantly (the accuracy is less than 7\%). This shows that there are still some challenges for the text-to-text models in general reasoning ability about the object affordances. In addition, we find that, the multi-modal models do not perform well in text-only settings, despite being pretrained on text and image data. The performances of the language models get boosted when ensembling with the multi-modal models, which indicates that the prediction of object affordance from sentence is a difficult task, and can be enhanced in presence of images. In addition to evaluating generative models, we establish a random baseline (Detailed in Appendix~\ref{appendix:random_baseline}).  Interestingly, we find that models like Flan-T5-large and Flan-T5-XL underperform compared to this random baseline in zero-shot settings.\\
\noindent\textbf{Finetuning on commonsense datasets.}
We observe that the fine-tuned model on commonsense reasoning task (Table~\ref{tab:commonsense_affordance_prediction}) show improved performance for the affordance prediction task. This indicates that the pre-trained models lack the reasoning of object affordance. Interestingly, we find that the smallest BERT-base model fine-tuned on PIQA, performs almost similar to that of the BERT-large or BART-large models (see Table~\ref{tab:zero_shot_affordance_prediction}).
\begin{table}[]\centering

\scriptsize
\begin{tabular}{lrr}\Xhline{2\arrayrulewidth}
\rowcolor{blue!10} \multicolumn{3}{c}{\textbf{MLM based}} \\\hline
\textbf{Model} &\textbf{Accuracy} &\textbf{mAP} \\
BERT-base-uncased-finetuned-piqa &0.45 &0.26 \\
BERT-large-uncased-finetuned-piqa &0.56 &0.29 \\
RoBERTa-large-finetuned-piqa &\textbf{0.64} &\textbf{0.45} \\
BART-large-finetuned-piqa &0.59 &0.35 \\
\bottomrule
\end{tabular}
\caption{\footnotesize Affordance prediction using models trained on commonsense data. Best results are marked in bold.}\label{tab:commonsense_affordance_prediction}
\vspace{-1.7\baselineskip}
\end{table}

\noindent\textbf{Few-shot performance.} We find that, in presence of few examples from our affordance dataset, the reasoning capability about object affordances can be enhanced for the PTLMs. The results with 124 shots (62 pairs as discussed earlier) are noted in Table~\ref{tab:zero_shot_affordance_prediction}. In Table~\ref{tab:affordance_prediction_generative}, we note the results for the in-context learning performance of the generative LLMs and VLMs. We observe a significant performance gain over zero-shot settings. Having said that, we also observe that, even with the in-context learning, the performance of the generative models (with more than 7b parameters) do not reach even close to the performance of the fine-tuned BERT-large model (340M parameters). This suggests that, for the specific affordance prediction tasks from text, finetuning is absolutely essential even for the state-of-the-art LLMs and VLMs.

\if{0}\begin{figure}[!htp]
    \centering
    \includegraphics[width=\columnwidth]{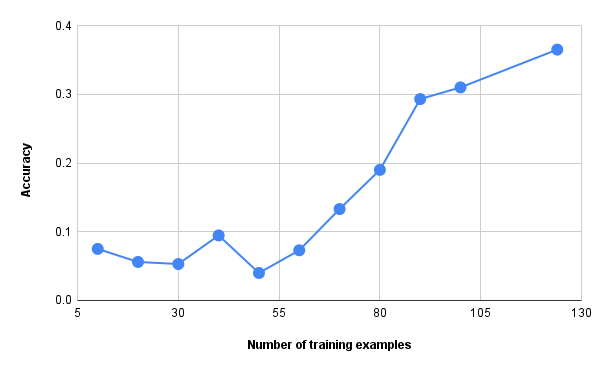}
    \caption{\footnotesize Few-shot performance growth of FLAN-T5-large.}
    \label{fig:few_shot_flan-t5}
\end{figure}\fi




\noindent\textbf{Error analysis}\\
\noindent\textit{Encoder based models.} We conducted a qualitative analysis of the erroneous cases for the two models (BART-large and RoBerta-large) in MNLI settings to understand what are the typical causes of errors. We take examples where accuracy is below 0.3. Consider the representative example below.

\begin{mdframed}[backgroundcolor=gray!10]
\textit{\underline{Sentence}: The salt from La Mata is often used as table salt.}
\textit{\underline{Object}: table salt}

\noindent
Top 5 predicted affordances (according to the probability score) - [`sitOn', `pourFrom', `grasp', `fix', `lookThrough']
\end{mdframed}

The model predicts `SitOn' as the top affordance for table salt, implying that the model misinterprets ``table salt'' with ``table''. Similarly, for the object ``the window sill'', the model predicts `lookThrough', `watch' as top affordances, which again suggests that the model is confused between ``the window sill'' and a ``window''. In another case, the model predicts ['grasp', 'writing', 'typing', 'lookThrough', 'throw'] as the top affordance labels for the object ``any rock concerts''. \\
\noindent \textit{Analysis of generative models.}~ 
  In Appendix Figure~\ref{fig:classwise_error_rate_vs_agreement_chatGPT}, we plot the correlation between error rate made by chatGPT for each affordance classes and the classwise annotator agreement. We observe a moderately negative correlation ($\rho = -0.29$) which suggests that there is a chance that the model is making higher mispredictions where the agreement is low. Similarly we observe that the mispredictions made by chatGPT for the most frequent objects has a moderately negative correlation ($\rho = -0.58$) with the annotator agreement. The correlation is shown in Figure~\ref{fig:most_frequent_objectwise_error_rate_vs_agreement_chatGPT}. The trends are similar for the other LLMs. These results together indicate that those objects and affordance classes which are hard to disambiguate by humans also pose a challenge to the most sophisticated GenAI models in predicting the correct answer. 

\section{\textsc{Text2Afford} for physical reasoning}
Apart from benchmarking LLMs and VLMs,  we observe whether Text2Afford can be used as a source of affordance knowledge. We choose the physical commonsense reasoning as a target as the
`Object affordance’ represents an innate physical property of an object, and we believe that any language model having strong affordance reasoning capability can enhance the physical reasoning capability.  To explore this, we perform an `instruction fine-tuning’ on the \textsc{Text2Afford} dataset (although it is not meant for training) using few open-source LLMs (\texttt{llama-3-8b-instruct}, \texttt{flan-t5}), and test on two physical reasoning dataset - (1) PROST \cite{aroca-ouellette-etal-2021-prost}, which contains 10 types of different physical properties of an object (including 6 affordance properties - rolling, breaking, stacking, grasping, sliding, bouncing) along with complex reasoning questions, and (2) PIQA \cite{Bisk2019PIQARA} which, focuses on selecting appropriate option given a situation that requires physical commonsense.\\
For PROST, using \texttt{llama-3} the accuracy boosts from 0.36 to 0.42 after instruct fine-tuning with \textsc{Text2Afford}. Moreover, out of the 6 affordance properties from PROST, the accuracy got boosted for the reasoning of 5 affordance properties. For the PIQA, the same LLM gives a maximum of 4\% accuracy boost. The full result is shown in Table~\ref{tab:physical_reasoning_evaluation_finetuned_on_text2afford}. This suggest the generalizability of \textsc{Text2Afford} in physical reasoning tasks.

\begin{table}[!htp]\centering
\renewcommand{\arraystretch}{1.1}
\resizebox{\columnwidth}{!}{
\begin{tabular}{l|
    >{\centering\arraybackslash}m{2cm}
    >{\centering\arraybackslash}m{5cm}|
    >{\centering\arraybackslash}m{2cm}
    >{\centering\arraybackslash}m{5cm}
}
    \hline
 \multirow{3}[1]{*}{\textbf{Model}} &\multicolumn{4}{c}{\textbf{Dataset}} \\\cline{2-5}
&\multicolumn{2}{c|}{\textbf{PROST}} &\multicolumn{2}{c}{\textbf{PIQA}} \\\cline{2-5}
&\textbf{Zero-shot} &\textbf{+\textsc{Text2Afford}} &\textbf{Zero-shot} &\textbf{+\textsc{Text2Afford}} \\\hline
\texttt{Llama-3-8b} &0.36 &0.42(\colorbox{green!30}{\textbf{+.06}})* &0.74 &0.78(\colorbox{green!30}{\textbf{+.04}})* \\
\texttt{FLAN-T5-xl} &0.13 &0.16(\colorbox{green!30}{\textbf{+.03}}) &0.57 &0.59(\colorbox{green!30}{\textbf{+.02}})\\
\texttt{FLAN-T5-xxl} &0.34 &0.38(\colorbox{green!30}{\textbf{+.04}})* &0.72 &0.75(\colorbox{green!30}{\textbf{+.03}})\\
\hline
\end{tabular}
}
\caption{\footnotesize Text-only physical reasoning dataset evaluation using different LLMs fine-tuned on \textsc{Text2Afford}. +\textsc{Text2Afford}: instruction fine-tuned on \textsc{Text2Afford}. * indicates $p$-value ($<0.05$) using Mann-Whitney U-Test.}\label{tab:physical_reasoning_evaluation_finetuned_on_text2afford}
\vspace{-1\baselineskip}
\end{table}

\section{Additional details}
\label{sec:additional_details}

\paragraph{Reason for choosing XNLI.} \textcolor{black}{We select XNLI to incorporate object references from less conventional and commonly explored scenarios. Unlike typical object identification datasets, XNLI offers sentences derived from novels, thus presenting a more in-the-wild textual context, which adds complexity and diversity to our dataset. Specifically, we choose the hypothesis portion of the XNLI sentences due to its shorter context length. This choice intentionally poses a challenge to LLMs, allowing us to better evaluate their reasoning capabilities, especially when dealing with minimal contextual information.} 

\paragraph{Non-explicit mention about contextual object affordance in the instruction.} \textcolor{black}{The instructions shown in Appendix Figure~\ref{fig:instruction} represent the initial guidance provided to annotators as an introduction to the task. Since understanding contextual object affordance can be challenging for non-expert annotators, this initial step was designed to give a basic idea of the task. However, we follow this up with a comprehensive training process and conduct two AMA (Ask Me Anything) sessions to ensure that annotators fully understood the need to base their judgments on the provided context. These efforts are key in ensuring high-quality annotations throughout the dataset creation.}
\paragraph{Reason for choosing 0-3 Likert scale in data annotation.} \textcolor{black}{We opt for a 0-3 Likert scale (4-point) to minimize the potential for neutral or non-committal responses, which can often arise when a midpoint option is available. Our initial observations indicated that some annotators tended to select an ``average" value without fully considering the contextual affordance of the objects, which diminished the depth of their evaluations and limited the discussion around ambiguities. By adopting a 4-point scale, we aim to encourage more decisive judgments. In addition, we provide a textbox (see Appendix Figure~\ref{fig:instruction}) for annotators to express any uncertainties or ambiguities they encountered, which has helped us in capturing more nuanced feedback.}
\paragraph{Reason for choosing visual genome.} \textcolor{black}{We chose visual genome as a primary source for real images due to its rich, complex scenes, which are widely used in visual reasoning tasks. The complexity of the images in visual genome provides diverse contexts that align well with the goals of our study, which focuses on contextual object affordances. While other methods, such as using search engines like Bing, have been employed in prior work to retrieve images, we opt for visual genome to ensure that the images contain sufficient contextual and visual detail to support affordance prediction, even if there are minor limitations in reasoning.}

\paragraph{Reason for choosing stable diffusion.}\textcolor{black}{Regarding the use of stable diffusion, we have been inspired by its demonstrated capability to generate high-quality, realistic images, particularly in prior studies where it was effective in reasoning tasks. While CLIP is primarily trained on real-world images, we hypothesize that stable diffusion could generate contextual images with sufficient accuracy to complement the real images. The generated images provide additional diversity, which helps us explore the affordance prediction task from a different angle. The benefit of using stable diffusion lies in its ability to create controlled, context-specific images that may not always be available in existing datasets, providing a broader range of testing scenarios for our models.}
\paragraph{Reason for framing generative tasks as a binary decision problem.} \textcolor{black}{In the generative setting, we opt for a binary yes/no classification to evaluate the affordance of individual context-object-affordance triples. We decide this based on the observation of the tendency of smaller LLMs to hallucinate, which can make direct affordance prediction challenging, particularly in zero-shot scenarios. By framing it as a binary classification task, we aim to simplify the evaluation and obtain more reliable results. In addition, our approach allows for a comprehensive evaluation of both positive and negative affordances. This is critical for our dataset, as it is designed to assess affordances that are applicable, as well as those that are not, in a given context.}

\section{Conclusion}
In this paper we introduced a novel text-based affordance dataset \textsc{Text2Afford} to investigate the affordance knowledge of PTLMs and pre-trained VLMs in different zero-shot settings. Our findings suggest that, the state-of-the-art language models, particularly text-to-text models, still exhibit limitations in their ability to reason about object affordances. In this seemingly easy task, we observe how context can introduce various levels of ambiguity and difficulty. We also observe, that even in the presence of such difficulty, human performance is superior and LLMs/VLMs still face difficulty in gaining such knowledge during their pretraining. Additionally, we observe how our dataset provides some additional knowledge that can be useful for physical commonsense reasoning -- stressing its orthogonality more with respect to the pretraining knowledge LLMs and VLMs possess.


\section*{Acknowledgments}
We would like to express our sincere gratitude to our co-authors for their invaluable contributions throughout this work. We also extend our thanks to the reviewers for their constructive feedback, which significantly helped improve the quality of the paper. Additionally, we gratefully acknowledge the support of the Toloka Research Grant program, which partially funded the data annotation process.

\section*{Limitations}
All of our experiments were conducted for English language. The models may act differently in multi-lingual settings. Our dataset is curated based on a specific set of affordance classes, which may introduce bias in terms of affordance representation. This could limit the generalizability of our findings to other domains or contexts. Despite efforts to train annotators and ensure agreement, subjective interpretations of affordance classes, can introduce noise. Our study primarily relies on textual information for affordance prediction. The absence of grounded visual information may limit the model's ability to accurately predict affordances, as some affordances may be more visually dependent.
\section*{Ethics Statement}
We used the publicly available XNLI corpus to curate our \textsc{Text2Afford} dataset. Our dataset does not contain any harmful or offensive contents. Any personal or sensitive information is anonymized and treated with utmost confidentiality. We ensure the protection of participants' privacy and obtain informed consent for data collection, annotation, and analysis. We incentivized all the annotators uniformly throughout the annotation process.




\bibliography{custom}
\bibliographystyle{acl_natbib}

\clearpage

\appendix

\section*{Appendices}

\section{Data annotation}

\subsection{Details of the \textsc{Text2Afford} dataset construction}
\label{appendix:dataset_construction_details}
\noindent\textbf{Preprocessing.} We select $20,000$ sentences from a crowdsourced English dataset (XNLI English) \cite{conneau2018xnli}\footnote{We choose XNLI as a source to facilitate multilingual extensions of our dataset.} and extract the noun phrases using the Stanford CoreNLP tool. As we restrict to the affordances that humans can directly perform, we filter the phrases which do not represent a tangible object (using ConceptNet). We manually filter out objects that cannot be acted upon directly by humans (such as school, building). After this preprocessing, we obtain a set of sentence-object pairs ($\langle x_i, o_i\rangle$), where the sentence acts as the context for the corresponding object. Each sentence on average has 2-3 such objects. 
We use the 15 predefined affordance classes from \citet{zhu2014reasoning} to label each sentence-object pair for annotation. 

We further expand our dataset with the labeled dataset provided by \citet{zhu2014reasoning}. Authors present 62 common objects and their corresponding 15 affordance labels. Given that our task is \textit{context-based affordance} prediction, we require to have sentence-object pairs for labelling. To generate diverse context for this dataset, we utilize the ChatGPT UI\footnote{\url{https://chat.openai.com}}\footnote{Prompt used: Can you make realistic sentences with the following objects? Followed by the list of object names.} model to generate synthetic sentences for each of the objects, followed by careful manual correction.\\
\noindent\textbf{Pilot studies \& annotator training.} We annotate the dataset using the Toloka platform\footnote{\url{https://toloka.ai/}}.  
We design an interface on this platform, which contained clear instructions and examples for annotating the data. We conduct two rounds of pilot studies to analyze the subjective understanding of the annotators and, thereby, filter out the high quality, serious annotators. For the \underline{first pilot study}, we present the annotators with the smaller 62 sentence-object pairs and ask them to label the instance with each affordance class on a scale of $0$ to $3$, indicating whether or not the affordance can be performed on the object. Here, 0-1 indicates that the affordance cannot be performed (high-low) and 2-3 indicates that the affordance can be performed low-high). 
We will further use these 62 synthetic sentence-object pairs for few-shot training. For quality control, we select the top 90\% of the available annotators in the platform, who are proficient in English, and use computers to complete the tasks\footnote{We exclude mobile-users as we believe the instructions may not appear clearly on mobile devices.}. A total of 15 annotators labelled the data, and all of them were incentivized uniformly. After the first pilot, we find that there is an extremely poor agreement among the annotators, and the overall precision is around 28\%. Therefore, we moved on to a \underline{second pilot} \underline{study}. Here, we use all the 62 sentence-object pairs from the previous study, along with 32 randomly selected sentence-object pairs from the XNLI data. We use the top 30\% of the annotators (based on the quality determined by the platform) available on the platform, while other criteria remained the same. We annotate 32 sentence-object pairs ourselves, and use all the labelled examples as \textit{control} data points to guide the annotators while labelling. A total of 114 annotators (including the 14 annotators from the first pilot study) participated in this version of the pilot study. We assign a specific skill to the annotators who attained more than $30\%$ precision and $30\%$ recall. In total, 48 annotators passed this criteria. 
Through initial pilot studies, we learnt that without grounded images, the task appears quite subjective to annotators. The main goal of the pilot studies have been to understand the annotators' quality, their comprehension of the task, and their preferences for incentives per task. We have also conducted two additional AMA (Ask Me Anything) sessions with interested annotators to further clarify the task.

\noindent\textbf{Final annotation.} In the final phase, we conduct the annotation on a larger set of sentence-object pairs, carefully selecting a total of 2,368 pairs. To ensure diverse perspectives and minimize bias, we engage 25 skilled annotators in this phase. Three annotators independently annotated each of the sentence-object pairs. Each annotator meticulously evaluated the affordance classes for every pair, contributing to a comprehensive annotation of the dataset. We perform the annotations in phases and complete the full task over \underline{10 phases}. \\
\noindent \textit{Reason for multiple annotation phases.} We intentionally consider relatively small number of data points for annotation in a single phase to make the review process easier. We carefully reviewed each annotation and provided feedback with guidance in case of mistakes. For instance, annotators initially got confused with the affordance `Watch' as human can \textit{watch} any visual objects. In another instance, some annotators asked whether `Throw' can be valid affordance for the object `Kittens' as humans can perform `Lift', `Throw' to the object `Kitten'. We discussed these types of ambiguities with the annotators after each phase. We measured class-wise agreement and average agreement across all classes after each annotation phase to ensure the quality of the annotations. 
The overall statistics for this \textit{currently} constructed dataset -- \textsc{Text2Afford} is in Table~\ref{tab:data_stat}. Throughout the data processing pipeline, we put scrupulous attention to the quality control, including the use of pilot studies, iterative annotation refinement, and manual filtering. These measures ensure that the dataset is comprehensive, accurate, aligned with the objectives of the study {and can be reliably reused in future}. Overall, our \textsc{Text2Afford} dataset consists of $2368$ sentence-object pairs having  $\sim 100k$ annotations (2368 $\times$ 15 $\times$ 3).

\

\subsection{Additional analysis on the datapoints by human}
\label{appendix:additional_data_analysis}
\textcolor{black}{
To further interpret the difficulty (or ambiguity) of the datapoints, we filter out the ``sentence-object-affordance” triples based on the percentage annotator agreement. We categorize the triples into 3 sections:
\begin{mdframed}
Agreement $>0.75$: Total 26,411 triples
0.4<Agreement $<0.75$: Total 7,084 triples
Agreement $<0.4$: Total 2,025 triples
\end{mdframed}
In general, the average agreement is higher for negative affordance classes than that of positive classes, which implies that it is easier for humans to tell which `affordance’ is not applicable to a particular object.\\ 
We employ three postgraduate students and provide them with the same set of instructions. We randomly sample 200 datapoints from the high agreement category (>0.75), and 200 samples from the low agreement category (<0.4) and ask to annotate independently. For the high agreement category scenario, we observe that in 86\%, 87\%, 91\% of the cases their answers aligned with the majority voted answers. 
For the low agreement category, in most of the cases they feel there is not enough information in the context to answer about affordance. In some cases, it was easier to tell the affordance of the object alone, but the context made it difficult to answer. For example: \\
\textit{Context: ``SCR systems are primarily made from tree branches , lime and sawdust .”
Can a human ``Sit On” tree branches?}\\
Without the context, it is easier to say ``Yes”.}

\subsection{Instruction page on the Toloka platform}
\label{sec:appendix_instruction}
\begin{figure*}[!ht]
    \centering
    \includegraphics[width=0.7\textwidth]{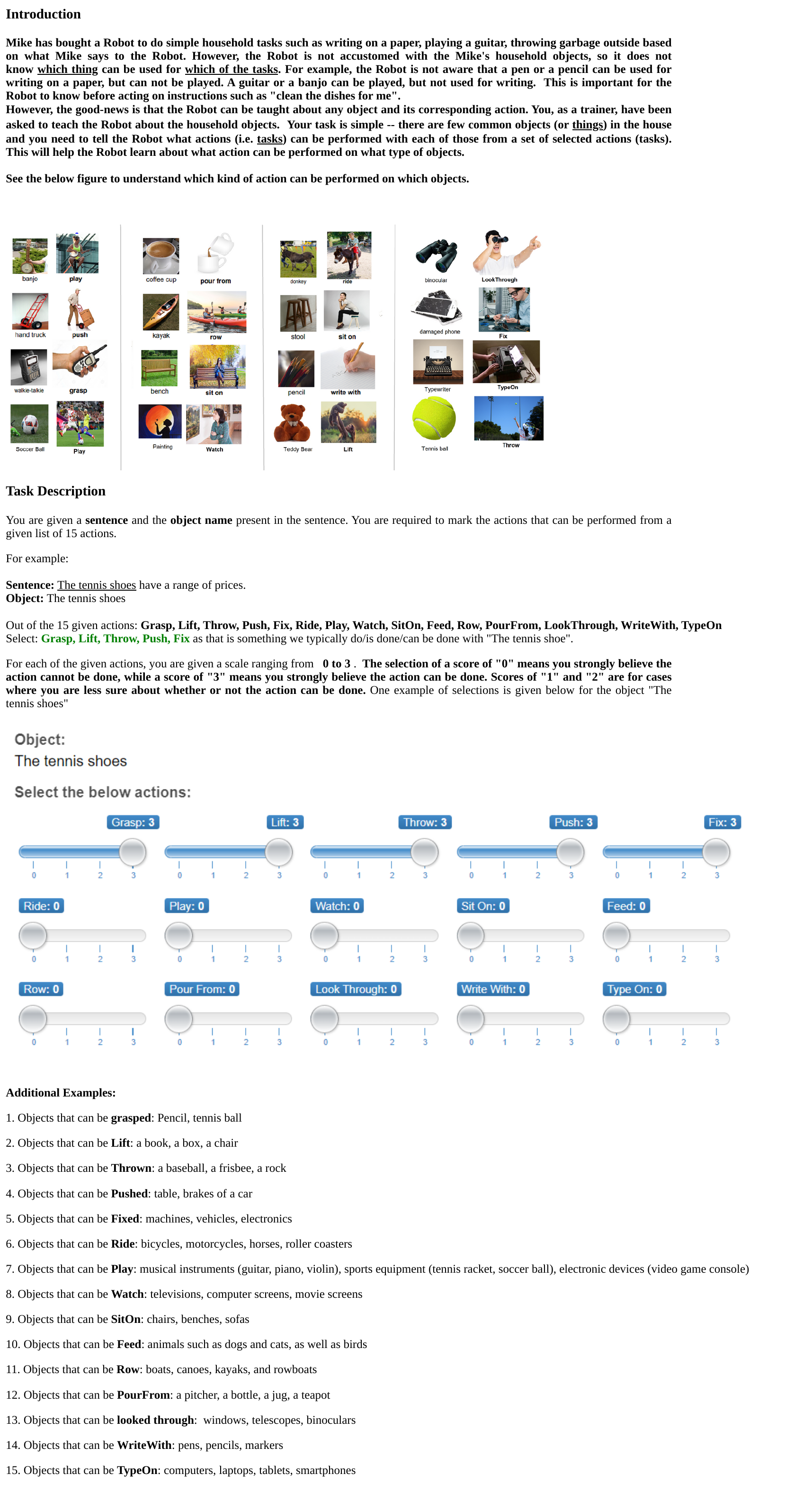}
    \caption{\footnotesize The instruction used for annotators in the Toloka platform}
    \label{fig:instruction}
\end{figure*}

Figure~\ref{fig:instruction} shows the guidelines/instructions, that the annotators had to follow for labelling.

\subsection{Interface for labelling}
\label{sec:appendix_interface}
\begin{figure}[!ht]
    \centering
    \includegraphics[width=\columnwidth]{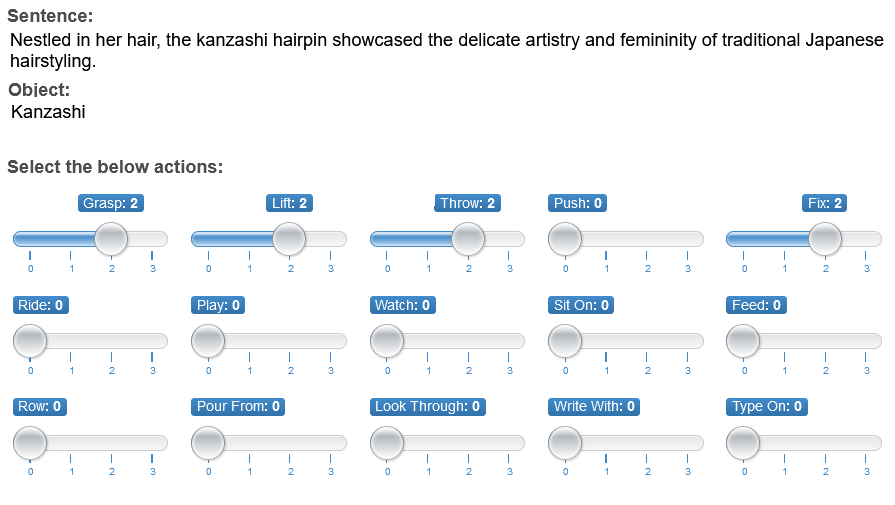}
    \caption{\footnotesize The sample task interface used for the annotators in the Toloka platform}
    
    \label{fig:interface}
\end{figure}

A sample task interface is shown in Figure~\ref{fig:interface}.

\subsection{Annotators demographics}
\begin{figure}[]
    \centering
    \begin{subfigure}{0.5\textwidth}
        \includegraphics[width=\linewidth]{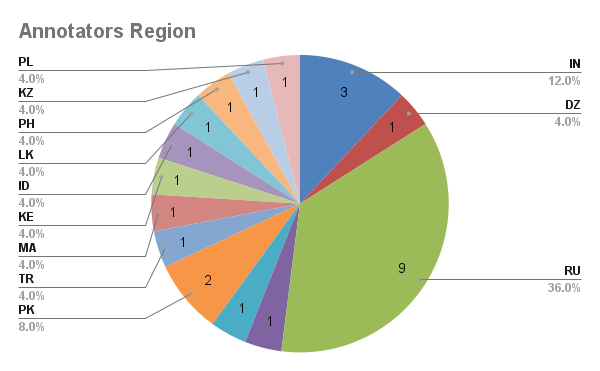}
        \caption{\footnotesize Country distribution of the annotators}
        \label{fig:annotator_region}
    \end{subfigure}%
    
    \hfill
    \begin{subfigure}{0.5\textwidth}
        \includegraphics[width=\linewidth]{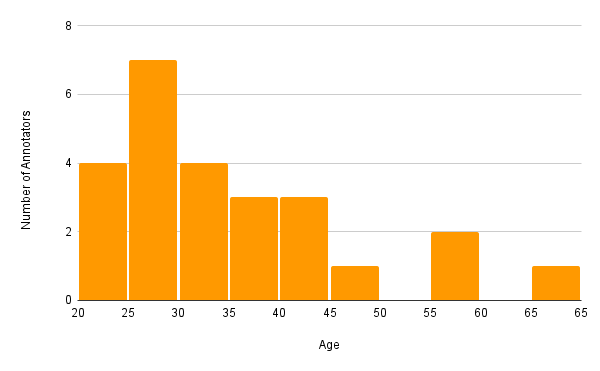}
        \caption{\footnotesize Age distributions of the annotators}
        \label{fig:age_distribution}
    \end{subfigure}
    
    \caption{\footnotesize The Annotators Demographics}
     
    \label{fig:annotator_demographics}
\end{figure}

Figure~\ref{fig:annotator_demographics} provides the demographic information about the annotators. We can observe that a large number of annotators (36\%) are from Russia and most of the annotators having the age in between 20-35.

\begin{figure*}[!thp]
    \centering
    \subfloat[]{\includegraphics[width=\columnwidth]{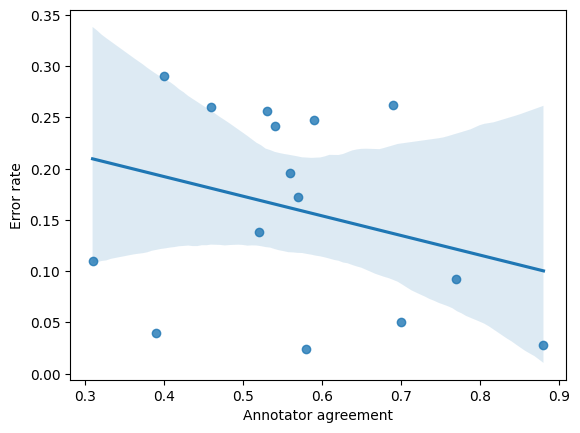}\label{fig:classwise_error_rate_vs_agreement_chatGPT}}
    \subfloat[]{\includegraphics[width=\columnwidth]{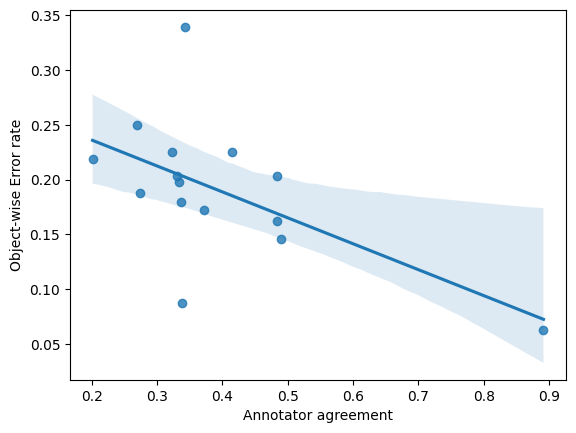}\label{fig:most_frequent_objectwise_error_rate_vs_agreement_chatGPT}}
    
    \caption{(a) \footnotesize Correlation between average classwise error rate made by chatGPT and the annotator agreement. ($\rho=-0.29$) (b) \footnotesize Correlation between frequent object wise error rate made by chatGPT and the annotator agreement. ($\rho=-0.58*$). *indicates a $p$-value < 0.05.}
\label{fig:correlation_chatgpt_combined}
\end{figure*}

\subsection{Phasewise annotator agreement}
\begin{figure}[htbp]
    \centering
    \includegraphics[width=\columnwidth]{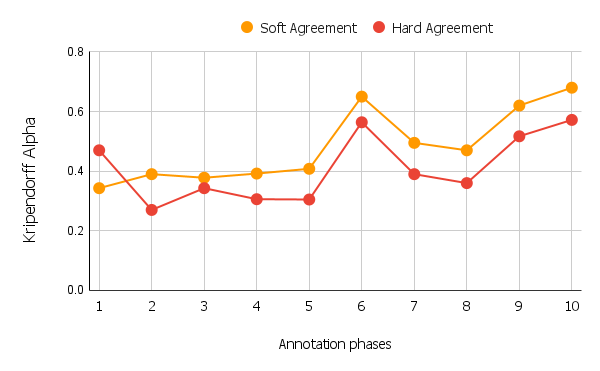}
    \caption{Phase-wise annotator agreement.}
    
\label{fig:phasewise_annotator_agreement}
\end{figure}

We plot the soft agreement\footnote{Soft agreement: Mapping Likert scale ratings to binary labels for measuring agreement by applying a threshold value.}, hard agreement\footnote{Hard agreement: Treating each Likert scale rating as a distinct label.} in Figure~\ref{fig:phasewise_annotator_agreement}, which shows gradual increase in agreement scores.

\subsection{Incentive details}
During the pilot study, we provided USD 0.05 per task-suite where in each task-suite, there were 10 examples (15 affordance labels for each example) to be answered. We attempted to take feedback from the tolokers who had answered randomly (e.g., mark all the values as 0), to understand their requirements properly. Most of them suggested that a wage of \$0.1 to \$0.15 would be ideal for the survey.

During the main study we provided USD 0.25 per task-suite, where in each task-suite there were 5 examples to be answered. Some of them were consistently providing good answers and few of them also suggested improvement on the objects. We awarded them with an additional bonus of USD 0.5. Overall, we spent USD 777 for the annotation process.

\begin{figure*}[!thp]
    \centering
    \subfloat[]{\includegraphics[width=\columnwidth]{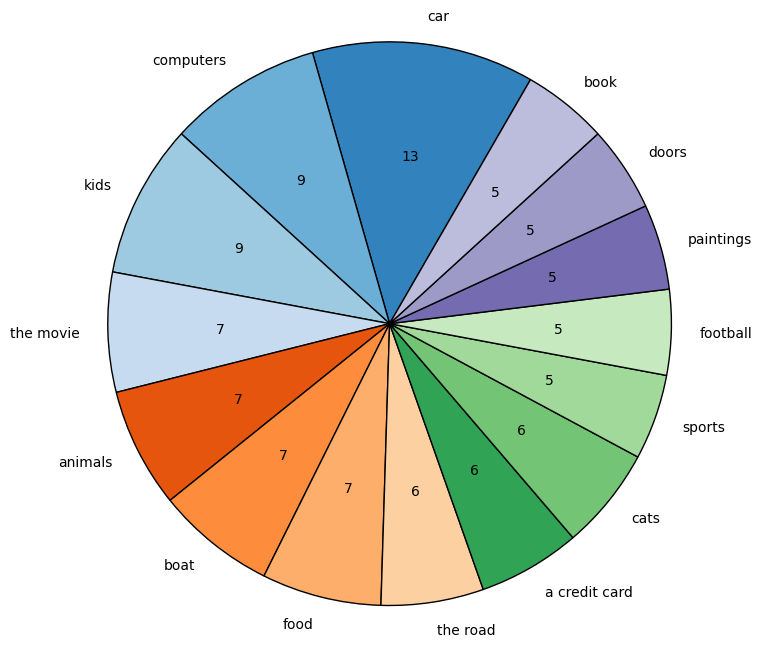}\label{fig:frequent_object}}
    \subfloat[]{\includegraphics[width=\columnwidth]{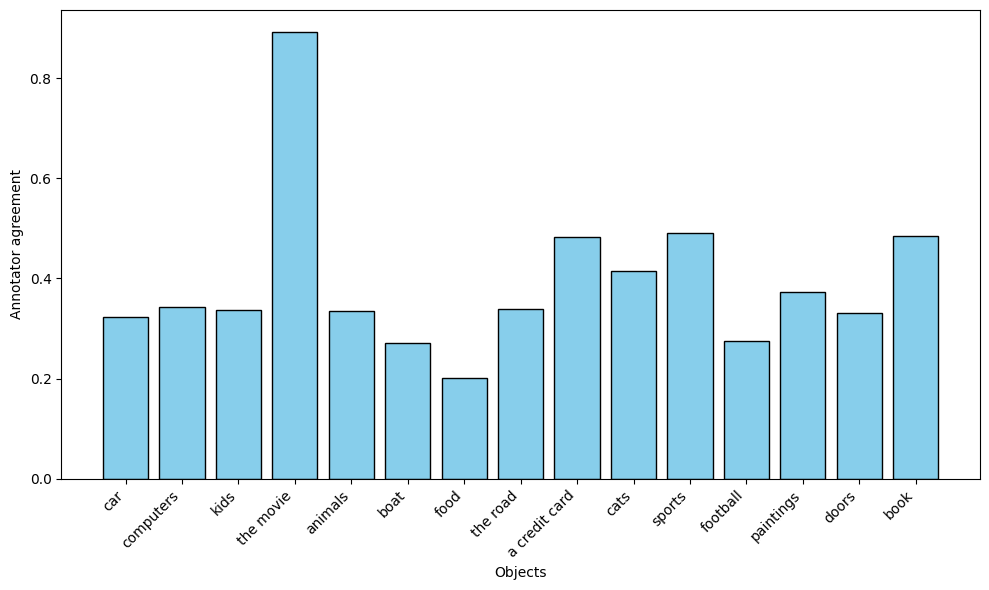}\label{fig:annotator_agreement_most_frequent_objects}}
    
    \caption{(a) Most frequent 15 objects and their corresponding frequency in the \textsc{Text2Afford} dataset. (b) Annotator agreement for the most frequent 15 objects.}
\label{fig:frequent_object_combined}
\end{figure*}

\subsection{Correlation of affordances}
\label{subsec:corr_afford}
\begin{figure}[!htp]
    \centering
    \includegraphics[width=\columnwidth]{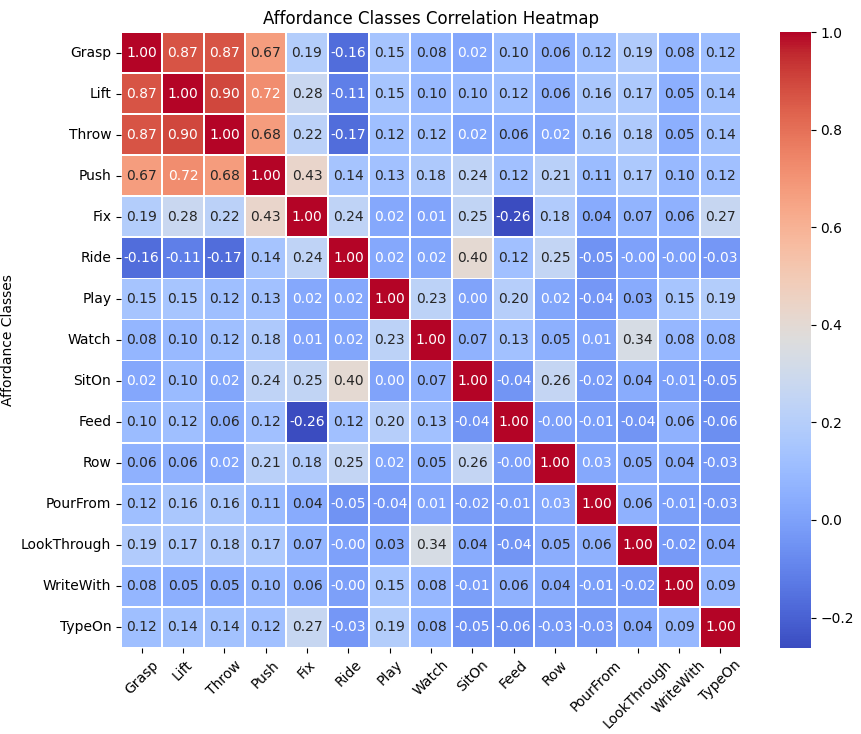}
    \caption{Correlation between each of the affordance classes.}
\label{fig:affordance_class_correlation}
\end{figure}

In Figure~\ref{fig:affordance_class_correlation} we show the correlation between the different affordance classes.

\subsection{Most frequent objects}
\label{subsec:most_frequent_objects}
Figure~\ref{fig:frequent_object} shows the most frequent 15 objects in the \textsc{Text2Afford} dataset.

\section{Experimental setup}
\label{appendix:exp_setup}

\subsection{Random baseline}
\label{appendix:random_baseline}
In addition to evaluating generative models, we establish a random baseline. For this baseline, we randomly assign "yes" to the 15 affordance classes for each sentence-object pair, with random selections made from 0 to 9 (based on the observation that the maximum number of positive affordances per pair is 9). Interestingly, we find that models like Flan-T5-large and Flan-T5-XL underperform compared to this random baseline in zero-shot settings, highlighting the inherent difficulty of the task in such scenarios.

\subsection{Zero-shot experiments}
\subsubsection{Commonsense reasoning tasks}
\label{appendix:zero-shot-commonsense-reasoning}
To understand whether the injection of the common
sense knowledge in the pre-trained models can enhance the performance of the affordance prediction, we first fine-tune the pre-trained models on common
sense reasoning dataset such as PIQA \cite{Bisk2019PIQARA}. Then we run the fine-tuned models on our dataset using the MLM setup. We use BERT-base, BERT-large, RoBERTa-large, and BART-large finetuned on MNLI.
\\
\noindent\textbf{NLI based approach.} 
The NLI task considers a premise and a hypothesis as input pair $\langle p,h\rangle$, and the models are trained to predict the probability whether the hypothesis is entailed by, contradicts or neutral with respect to the premise. Here we use the entailment probability from the models: $p_{La}(h|p) = p(l=``ENTAILMENT" | (p, h)).$
This approach requires language models to be fine-tuned on premise-hypothesis pairs with the corresponding labels. Here we use RoBERTa-large and BART-large fine-tuned on the Multi-genre NLI (MNLI) corpus \cite{williams-etal-2018-broad} consisting of 433k sentence pairs. For each sentence-object pair in our dataset as the premise, and use the hypothesis as ``\textit{<object> can be used for <affordance> by human}'' for each object present in the sentence and 15 affordance classes. Using the NLI setting, we predict the entailment score for each affordance class for the given sentence-object pair. We use these scores for ranking the affordance classes and report mAP scores as well as accuracy. 

\subsection{Ensemble language and vision prediction} 
\label{appendix:exp_ensemble_VL}
Following \citet{yang-etal-2022-z}, we use the weighted sum as the late fusion over the final output probabilities of each affordance class from the language and multi-modal models. Before late fusion, we normalize the output probability scores from different models.  We calculate the score as: $P_{ens}(y|x) = (1-w)p_{L_a}(y|x) + wp_{V_I}(y|x)$ where $w$ is the relative size of the vision-text model and the language model (following \citet{yang-etal-2022-z}):
$w = Sigmoid\left(\frac{\rho_{V_I}}{\rho_{L_a}}\right)$.
Here $\rho_{V_I}$ and $\rho_{L_a}$ denote the number of parameters
of the multi-modal and language models respectively.

\subsection{Few-shot experiments}
\label{appendix:few-shot-training-data}
\paragraph{Training data} To create few-shot training examples for fine-tuning encoder based PTLMs, we take all the 62 objects, and for each object we randomly select exactly 1 positive affordance class (i.e., the class label annotated as 1) and 1 negative affordance class (i.e., the class label annotated as 0) for generating the training prompt. As this dataset does not contain any context sentences for a corresponding object, we use ChatGPT UI to generate the sentences for the corresponding objects and manually verify the sentences, so that it does not contain any invalid information. Finally, we have 62 sentence-object pairs and 2 classes (one positive and one negative) per pair, which we use to generate training examples. Each training example consists of a prompt and a label. They constitute 124 training examples (62 sentence-object pairs and 2 selected classes for each) for the few-shot experiment. \\
\paragraph{Selecting examples for in-context learning}: We randomly sample five sentence-object-affordance triples from the above training data as the incontext demonstration examples in such a way that there should be $k$ positive affordance classes. We vary the number of positive affordance classes $k \in \{1, 2, 3\}$ and report the average accuracy. \\
\paragraph{Experimental setup.} \label{appendix:few-shot-experimental_setup} We fine-tune the encoder based language models using the training data, and for the generative LLMs and the VLMs, we utilize the training data to select in-context demonstration examples.\\
\noindent\textit{Fine-tuning PTLM:} We fine-tune the PTLMs in two different setups - NLI based and prompt based. For the NLI based setup we have the context sentence as premise and use same prompt (i.e., ``<object> can be used for <affordance> by human'') which we use in the zero-shot settings as hypothesis. We use label as 1 for the positive affordance and label as 0 for the negative affordance. We use BERT-large-uncased, RoBERTa-large and BART-large for fine-tuning in this setup. We reuse these fine-tuned models for few-shot predictions in MLM setup. We use Adam optimizer with a learning rate of $2\times 10^{-5}$. We fine-tune the model for 5 epochs for each case.\\
\noindent\textit{In-context learning for generative models}: We employ the same generative LLMs as well as VLMs to perform affordance prediction using \textit{five} demonstration examples from the training data. We use the same text prompt as zero-shot setting and concatenate the five demonstration examples along with corresponding label (i.e., `YES' for positive class, and `NO' for the negative class) to the prompt and ask the LLMs and VLMs to predict the affordance. In case of the VLMs, we do not provide any additional image example here.

\subsubsection{Multimodal task setup}
\label{appendix:multimodal-task-setup}
Images contain necessary information about shape, texture, and size of objects that can be utilized to effectively predict an object affordance (such as the handle of the bucket can be used to grasp and lift). Hence, we also convert the problem into a multi-modal task by retrieving (or generating) a corresponding image from the context sentence, and  predict the affordance of an object (mentioned in the sentence) based on the input.\\ 
\noindent\textbf{Synthesizing images.}  In this setup, we use two different techniques to synthesize \textit{semantically close} images to corresponding context sentences using 1) retrieval and 2) generation. We further use top five images for both, to get an accurate estimation. \\
\noindent\textit{Retrieval based}: We employ Visualgenome \cite{krishna2017visual} dataset, consisting of 108,077 images and ~3.8 million object instances as the image database. We first encode the images using multi-modal CLIP \cite{radford2021learning} based sentence-transformers architecture, and index those image embeddings using Approximate Nearest Neighbour search (ANN)\footnote{\url{https://pypi.org/project/annoy/}}, for making the search efficient. Now, for each sentence, we search for top five images from the database to be used further.\\
\noindent\textit{Generation based}: Recently, the multi-modal generative models  \cite{ramesh2022hierarchical,saharia2022photorealistic} have shown incredibly good performance for text based image generation tasks. We adopt the recent \textit{StableDiffusion} \cite{rombach2022high} model to generate top five images based on the sentence as a text prompt.\\

We use the top five retrieved images by using retrieval and generation methods each. We use \textit{CLIP} \cite{radford2021learning} and \textit{ViLT} \cite{kim2021vilt} as our vision-text models. CLIP is
pre-trained on 400M image-caption pairs with the contrastive learning strategy. CLIP has a text encoder $f_T$ and a visual encoder $f_V$, which can project text and image into the shared latent space. We aggregate the $k$ (=5) corresponding images and use CLIP to compute the relevance score of (x, y): $Score_{VI}(x, y) = \frac{1}{k}  \sum_{i=1}^{K}{\cos{(f_T(x), f_v(I^k_y))}}$,
where $I^k_y$ is the $k^\textrm{th}$ image for the input text $y$. 
In the ViLT model we provide the text prompt along with the representative images as input to predict the masked token. We use the same prompt as the previous MLM task (i.e., ``\textit{<Object> can be used for <MASK\_TOKEN> by human.}'') and get the probability of each affordance class as the logit corresponding to the <MASK\_TOKEN>.\\
\noindent\textbf{Text generation based. } Similar to section~\ref{exp:generative_LLM}, we utilize state-of-the-art VLMs to make predictions regarding object affordances. We provide with a `YES\textbackslash NO` question answering based text prompt along with the aligned images as input to the VLMs, and the model should generate an answer whether a particular affordance can be performed on the given object. We use state-of-the-art VLMs such as IDEFICS \cite{laurenccon2023obelisc}, LLaVA \cite{liu2023llava}, InstructBLIP \cite{instructblip} for this task. The text prompt used for the models can be found in the Appendix~\ref{appendix:prompts}, Table~\ref{tab:prompts}.

\section{Additional (mis)prediction analysis}
\subsection{Affordance classwise mis-prediction}
\textcolor{black}{We analyze the mis-prediction rates for each class using the best LLMs (chatGPT, llama-3-8b). We observe that, the classwise mis-prediction rate is similar to the distribution of each class in the original data, i.e., the classes such as ‘grasp’, ‘lift’ having higher mis-predictions compared to ‘typeOn’, ‘row’.} 

\subsection{Objects with multiple positive affordances}
\textcolor{black}{We conduct an analysis to determine whether the frequency of positive affordances for an object impacts model accuracy. Our findings indicate that the accuracy is highest when an object has a single positive affordance. Beyond this point, the number of positive affordances does not significantly influence the model's performance. Specifically, we observe that as the number of positive affordances increases, the accuracy fluctuates without a clear pattern, suggesting that additional positive affordances do not contribute to a consistent improvement or decline in model accuracy.}

\subsection{Correlation of ChatGPT accuracy and average human agreement}
We provide the figures corresponding to the generative model analysis in Figure \ref{fig:correlation_chatgpt_combined}.

\section{Prompt selection}
\label{appendix:prompts}
We use intuitive prompts for each of the setups, which are suitable for affordance related to object.  

\begin{table}[!htp]\centering
\scriptsize
\begin{tabular}{l|p{20em}}\toprule
\textbf{Model} &\textbf{Prompt used} \\\midrule
FLAN-T5 &consider \{sentence\}. Now, from this information can human \{affordance\} the \{object\_name\}? Answer YES or NO: \\\hline
Falcon &"""You are a helpful AI assistant. Answer only "YES" or "NO" for the question based on the given context. 
Context:{sentence} {\textbackslash}n
>>QUESTION<< Can human \{affordance\} the \{object\_name\}? {\textbackslash}n
>>ANSWER<<""".strip() \\\hline
I-BLIP, IDEFICS, LLaVA &consider the sentence \{sentence\}. Now from this information, can human \{affordance\} the \{object\_name\}? Accompanying this query is an image of the {object\_name}. Note that the image may contain noise or variations in appearance. Given the textual description and the image, answer YES or NO whether the human can \{affordance\} the \{object\_name\}. Answer: " \\
\bottomrule
\end{tabular}
\caption{Prompt format used by different models for the prediction. I-BLIP: InstructBLIP.}\label{tab:prompts}
\end{table}

\section{Instruction fine-tuning setup}
\label{instruction finetuning setup}
\paragraph{Data sample selection.} \textcolor{black}{We select sentence-object pairs from the \textsc{Text2Afford} dataset where at least one positive affordance is present. For each selected sentence-object pair, we randomly assign one positive affordance and one negative affordance, yielding a balanced dataset of 1819 training instances (positive and negative classes). To incorporate additional domain knowledge and reduce the likelihood of generating hallucinated answers, we include 500 randomly sampled instances from the training set of the target task (i.e., PIQA). For the PROST task, as the training set is not explicitly available, we sample from the test set and ensure these samples are removed from the evaluation set during testing. The training instances are framed in a multiple-choice question answering format.}
\paragraph{Fine-tuning setup.}\textcolor{black}{We utilize Alpaca-formatted prompts (shown in Table~\ref{tab:instruction_text2afford}, Table~\ref{tab:instruction_PIQA} and Table~\ref{tab:instruction_PROST} for the \textsc{Text2Afford}, PIQA and PROST tasks, respectively). We fine-tune 4-bit quantized models with PEFT, focusing on the adapter layers. We perform the fine-tuning over 5 epochs with a batch size of 8, a learning rate of 2e-10, weight decay, and a maximum sequence length of 256.}

\begin{table*}
\tcbset{
  fonttitle=\bfseries,
  boxrule=0.5mm,
  width=\textwidth,
  arc=4mm,
  auto outer arc,
  boxsep=2mm,
}

\begin{tcolorbox}[title=Instruction to fine-tune \textsc{Text2Afford}]

Below is an instruction that describes a task. Write a response that appropriately completes the request.\\

\noindent \textbf{\#\#\# Instruction:}\\
You are an AI assistant that has strong reasoning capability. You are given a context containing an object, and you are asked to answer a question about the object based on the context. Just response 'Yes' or 'No'.\\ 

\noindent \textbf{\#\#\# Context:}\\
\{context\}\\

\noindent \textbf{\#\#\# Object:}\\
\{object\}\\

\noindent \textbf{\#\#\# Question:}\\
Can human \{affordance\} the \{object\}?\\

\noindent \textbf{\#\#\# Answer:}\\
\{answer\}

\end{tcolorbox}
\caption{\label{tab:instruction_text2afford} \footnotesize Instruction to fine-tune \textsc{Text2Afford.}}
\end{table*}

\begin{table*}
\tcbset{
  fonttitle=\bfseries,
  boxrule=0.5mm,
  width=\textwidth,
  arc=4mm,
  auto outer arc,
  boxsep=2mm,
}

\begin{tcolorbox}[title=Instruction to fine-tune PIQA]

Below is an instruction that describes a task. Write a response that appropriately completes the request.\\

\noindent \textbf{\#\#\# Instruction:}\\
You are an AI assistant that has strong reasoning capability. You are given a situation and asked to choose the most appropriate option from given two options.\\ 

\noindent \textbf{\#\#\# Situation:}\\
\{situation\}\\

\noindent \textbf{\#\#\# Options:}\\
\text{[0]} \{option0\}\\
\text{[1]} \{option1\}\\

Only response the  `answer id'. For example if the answer is [0] then response 0. DO NOT respond anything other than <0, 1>.\\

\noindent \textbf{\#\#\# Answer:}\\
\{answer\}

\end{tcolorbox}
\caption{\label{tab:instruction_PIQA} \footnotesize Instruction to fine-tune PIQA.}
\end{table*}

\begin{table*}
\tcbset{
  fonttitle=\bfseries,
  boxrule=0.5mm,
  width=\textwidth,
  arc=4mm,
  auto outer arc,
  boxsep=2mm,
}

\begin{tcolorbox}[title=Instruction to fine-tune PROST]

Below is an instruction that describes a task. Write a response that appropriately completes the request.\\

\noindent \textbf{\#\#\# Instruction:}\\
You are an AI assistant that has strong reasoning capability. You are given a question with 4 options and you have to choose the right option.\\ 

\noindent \textbf{\#\#\# Question:}\\
\{question\}\\

\noindent \textbf{\#\#\# Options:}\\
\text{[0]} \{option\_A\}\\
\text{[1]} \{option\_B\}\\
\text{[2]} \{option\_C\}\\
\text{[3]} \{option\_D\}\\

Only response the `answer id'. For example if the answer is [0] then response 0. DO NOT respond anything other than <0, 1, 2, 3>.\\

\noindent \textbf{\#\#\# Answer:}\\
\{answer\}

\end{tcolorbox}
\caption{\label{tab:instruction_PROST} \footnotesize Instruction to fine-tune PROST.}
\end{table*}

\section{Model implementation details}
The language models and the ViLT are built on top of the huggingface API\footnote{\url{https://huggingface.co/}}. For NLI based zero-shot prediction, we use the zero-shot classification pipeline \footnote{\url{https://huggingface.co/docs/transformers/main_classes/pipelines}}. We adapted the CLIP model from the OpenAI's public repo \footnote{\url{https://github.com/openai/CLIP}}, and we select the ViT/B32 as the image encoder. For ViLT, we select the vilt-b32-mlm \footnote{dandelin/vilt-b32-mlm} model. For generative LLMs and VLMs we apply the models available on huggingface \footnote{\url{https://huggingface.co/models}}. All the experiments were conducted on 2x NVIDIA RTX 4090 GPU server.

\section{Details of evaluation metric}
\label{appedix:evaluation_metric_details}
\textcolor{black}{
For a `Sentence-Object’ pair we calculate accuracy in the following way.
In the ground-truth, each affordance class is treated as a binary value, where a value of 1 represents a `positive affordance’ indicating that the affordance can be performed on the object, and a value of 0 represents a `negative affordance’ indicating that the affordance cannot be performed.
Now, for a particular `Sentence-Object’ pair, let’s assume there are two positive affordances (P1, P2) in the ground truth; then there will be 13 negative affordances (as we have a total 15 affordance classes).
In case of encoder-based models, for each positive affordance, we compare its prediction score against each negative affordance's score. If a positive affordance's score is higher, we increase the Correct count; otherwise, the Wrong count. Accuracy is calculated as Correct / (Correct + Wrong).\\
\noindent In case of encoder-decoder or decoder-only models, Due to the inherent difficulty in automatic evaluation, we predict 'YES\textbackslash NO' for each affordance class, mapping 'YES' to 1 and 'NO' to 0. Accuracy is then measured in the same way as for encoder-based models (assuming 1 or 0 as the score for each affordance class).}

\section{Dataset creation time}
Annotating affordances about the object from a text itself is a difficult and very subjective task. It took approximately 5 months for completing the extraction of noun-phrases from xnli data, filtering objects,  selecting skillful tolokers and training, and then final phase-wise annotation after rigorous review process. 

\section{Sample dataset}
\begin{figure*}[!t]
    \centering
    \includegraphics[width=\textwidth]{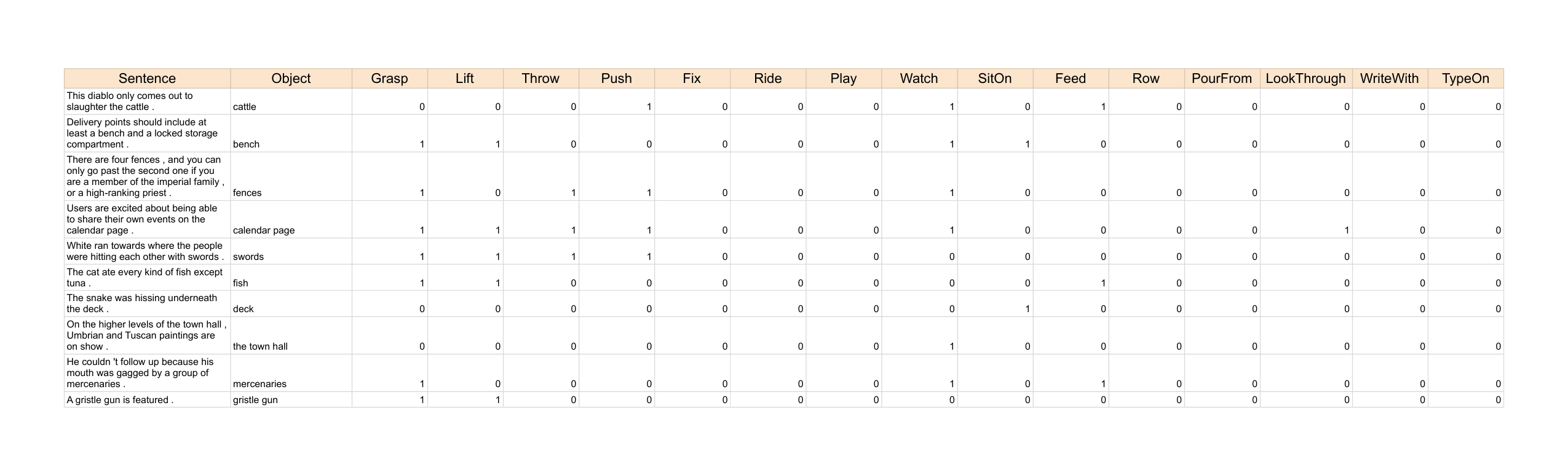}
    \caption{Example snapshot of \textsc{Text2Afford} dataset.}
    \label{fig:example_data}
\end{figure*}

Figure~\ref{fig:example_data} shows a sample of \textsc{Text2Afford} dataset

\section{Additional experiments}
\subsection{Qualitative analysis of generated images}
We conducted a qualitative analysis on 50 randomly sampled objects and their corresponding generated images. Two annotators (one Phd student and one undergrad student) marked each of the 5 generated images as 1 or 0 according to their relevance and non-relevance to the object respectively. We considered the image as relevant if both of the annotators marked that image as 1. We achieved an Acc@1 of 0.2, Acc@5 of 0.88 and an MAP@5 of 0.36. Which suggests that in most of the cases there are relevant images in the top-5 generated images. 
In our pursuit of assessing the statistical significance of our sampled data (i.e., the 50 examples), we embarked upon a rigorous hypothesis testing procedure utilizing the binomial distribution. Within our specific context, we accorded greater significance to the top-5 accuracy metric, which demonstrated an impressive achievement of 0.88. This signifies that among the 50 selected examples, in 44 instances, at least one of the five generated images displayed relevance to the object under consideration.

Guided by this success rate, we proceeded to conduct a meticulous hypothesis test employing the binomial distribution. We assumed an expectation of success at 0.75. The outcome of this statistical analysis revealed a p-value of less than 0.02, thereby underscoring the statistical significance of our success rate.

    

\end{document}